\pdfoutput=1
\documentclass[11pt]{article}

\usepackage[final]{acl}

\usepackage{times}
\usepackage{latexsym}

\usepackage[T1]{fontenc}
\usepackage[utf8]{inputenc}

\usepackage{microtype}

\usepackage{inconsolata}

\usepackage{hyperref}
\usepackage{url}
\usepackage{graphicx}
\usepackage{multirow}

\usepackage{subcaption}
\usepackage{cancel}

\usepackage{algorithm}
\usepackage{algpseudocode}

\usepackage{wrapfig}

\usepackage[table,xcdraw]{xcolor}

\title{\ourmethod: An Iterative and Integrated Framework for Verifiable Causal Discovery in the Absence of Tabular Data}

\author{
 \textbf{Tao Feng\textsuperscript{1}},
 \textbf{Lizhen Qu\textsuperscript{1} \thanks{Corresponding author}},
 \textbf{Niket Tandon\textsuperscript{2}},
 \textbf{Gholamreza Haffari\textsuperscript{1}}
\\
 \textsuperscript{1}Monash University,
 \textsuperscript{2}Microsoft Research, India
\\
 \small{
   \textsuperscript{1} {{firstname.lastname}@monash.edu},
   \textsuperscript{2} {nikett@gmail.com}
 }
}

\usepackage{amsthm,amsmath,amsfonts,bm,xspace}
\usepackage{color}

\def\eg{{\em e.g.,}\xspace}
\def\ie{{\em i.e.,}\xspace}

\def\Figref#1{Figure~\ref{#1}}

\def\eqref#1{equation~\ref{#1}}
\def\Eqref#1{Equation~\ref{#1}}
\def\Algref#1{Algorithm~\ref{#1}}

\def\1{\bm{1}}

\def\rz{{\textnormal{z}}}

\def\vr{{\bm{r}}}

\def\mM{{\bm{M}}}

\def\mV{{\bm{V}}}

\DeclareMathAlphabet{\mathsfit}{\encodingdefault}{\sfdefault}{m}{sl}
\SetMathAlphabet{\mathsfit}{bold}{\encodingdefault}{\sfdefault}{bx}{n}

\def\gG{{\mathcal{G}}}

\def\sD{{\mathbb{D}}}
\def\sR{{\mathbb{R}}}
\def\sS{{\mathbb{S}}}

\def\sX{{\mathbb{X}}}

\def\sZ{{\mathbb{Z}}}

\newcommand{\system}[1]{\text{#1}}
\newcommand{\ourmethod}{\system{IRIS}\xspace}

\begin{document}
\maketitle
\begin{abstract}
Causal discovery is fundamental to scientific research, yet traditional statistical algorithms face significant challenges, including expensive data collection, redundant computation for known relations, and unrealistic assumptions. While recent LLM-based methods excel at identifying commonly known causal relations, they fail to uncover novel relations. We introduce \ourmethod (\textbf{I}terative \textbf{R}etrieval and \textbf{I}ntegrated \textbf{S}ystem for Real-Time Causal Discovery), a novel framework that addresses these limitations. Starting with a set of initial variables, \ourmethod automatically collects relevant documents, extracts variables, and uncovers causal relations. Our hybrid causal discovery method combines statistical algorithms and LLM-based methods to discover known and novel causal relations. In addition to causal discovery on initial variables, the missing variable proposal component of \ourmethod identifies and incorporates missing variables to expand the causal graphs. Our approach enables real-time causal discovery from only a set of initial variables without requiring pre-existing datasets.\footnote{Our code and data are available at \url{https://github.com/WilliamsToTo/iris}}
\end{abstract}

\section{Introduction}
\label{sec:introduction}
\begin{figure*}[ht]
    \centering
    \includegraphics[width=0.9\linewidth]{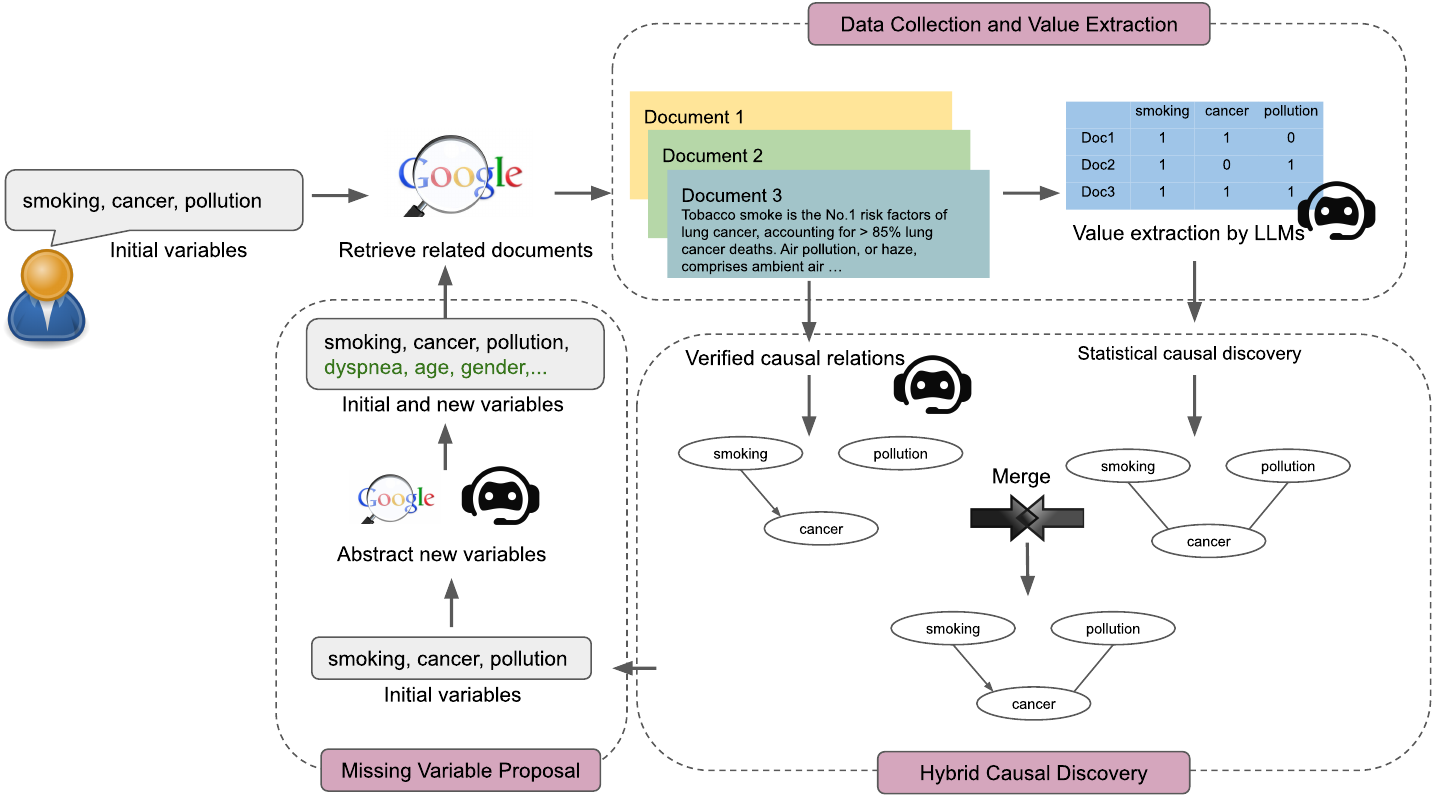}
    \caption{Illustration of \ourmethod. Given initial variables, we use the Google Search API and LLMs to \emph{collect relevant documents and extract variable values}, then form structured data. For \emph{hybrid causal discovery}, the statistical branch uses the structured data, while the causal relation extraction branch uses the retrieved documents. Their results are merged into the final causal graph. The \emph{missing variable proposal} component identifies new variables, which are iteratively fed into our framework to expand the causal graphs.}
    \label{fig:illustration_of_framework}
 % \vspace{-15pt}
\end{figure*}

A fundamental task in various disciplines of science, including biology, economics and healthcare, is to identify and utilize underlying causal relations \cite{Kuhn1962}. Although interventional experiments are ideal for discovering causal relations, they are often impractical due to ethical, financial, or logistical constraints. Therefore, researchers develop statistical methods to infer causal relations from purely observational tabular data \cite{pearl2009causality, spirtes2000causation}, though such data is often \textit{not} available for a wide range of NLP applications. 

Statistical and large language model (LLM)-based causal discovery algorithms face distinct challenges that limit their applicability in real-world scenarios.
First, traditional statistical algorithms predominantly require high-quality structured tabular data, which is notoriously difficult to obtain. In contrast, LLM-based methods can consistently estimate causal relations explicitly present in their training data without relying on tabular data. However, these models encounter significant limitations when attempting to uncover causal relationships that were not previously documented \cite{feng2024pretrainingcorporalargelanguage, ze2023causal}. %As a result, researchers frequently resort to synthetic data, limiting the generalizability of their findings \cite{dong2023versatilecausaldiscoveryframework, gasse2021causalreinforcementlearningusing, 10.5555/1941985, binder1997adaptive}. 
Second, statistical causal discovery algorithms require predefined sets of random variables as input, a constraint that significantly limits their flexibility. LLMs, however, demonstrate the capability to reliably extract and identify concepts and entities as variables directly from texts \cite{10.5555/3020548.3020641, 10.3389/fgene.2019.00524}. Third, most statistical algorithms are theoretically grounded and mathematically verifiable, but operate under assumptions that rarely hold in real-world scenarios, such as the \emph{causal sufficiency} assumption (\ie the absence of unobservable variables in the causal graph) and \emph{acyclicity} assumption (\ie the absence of cycles in the causal graph) \cite{pearl2009causality,neal2020causalitybook}. In contrast, the verification of LLMs' predictions in causal discovery remains an open challenge.

To address these limitations, we propose \ourmethod, \textbf{I}terative \textbf{R}etrieval and \textbf{I}ntegrated \textbf{S}ystem for verifiable causal discovery, in the absence of tabular data for statistical methods. To leverage the strengths of both statistical methods and LLMs, our framework takes a \emph{hybrid} causal discovery approach, combining statistical methods with LLM-based causal relation extraction and verification techniques. This hybrid strategy allows us to leverage known causal relations and uncovering novel causal relations.
\ourmethod begins with a set of initial random variables, which are sent as queries to retrieve a collection of relevant documents. Consequently, LLMs are applied to map the unstructured texts into structured tabular data, which is utilized by an appropriate statistical method to perform causal discovery. Its results are further merged with the causal relations predicted and verified by LLMs. 
  This hybrid approach allows cycles in causal graphs, thereby relaxing the \emph{acyclicity} assumption. Additionally, we introduce a variable proposal component to identify new variables that have causal relations with the initial variables. This component allows us to relax the \textit{causal sufficiency} assumption. We then iteratively use the expanded variables as input to our framework, further expanding the causal graphs.

Our experimental results demonstrate that \ourmethod significantly surpasses strong baselines across all datasets and scales effectively from small (4 initial variables) to large causal graphs (27 initial variables), as detailed in Section~\ref{sec:eval_expanded_causal_graphs}. Evaluations of individual components reveal that each component outperforms its corresponding baselines. Specifically, the evaluation of value extraction component shows that \ourmethod with GPT-4o exceeds the strong baselines, which also utilizes GPT-4o (Section~\ref{sec:eval_value_extraction}). Our hybrid causal discovery method consistently outperforms both statistical algorithms and LLM-based approaches (Section~\ref{sec:eval_causal_discovery}). Lastly, our variable proposal component is more effective compared to prompt-based baselines (Section~\ref{sec:eval_missing_variable_proposal}).

Primary contributions of \ourmethod are as follows: 1) We introduce an automatic sample collection and value extraction component that significantly reduces the manual labor for data collection in causal discovery tasks. 2) We propose a hybrid causal discovery method that leverages existing causal relations and uncovers novel causal relations. Our method permits cycles in causal graphs, thus relaxing the \emph{acyclicity} assumption. 3) We develop a missing variable proposal component that identifies new variables that may have causal relations with the initial variables, relaxing the \emph{causal sufficiency} assumption. 4) Experimental results demonstrate that \ourmethod consistently outperforms its baselines, with each component of \ourmethod also surpassing corresponding baselines.

\section{Background}
\label{sec:background}
Causal discovery focuses on uncovering causal relations within a set of variables. Given a pair of variables $(X, Y)$, the objective is to determine whether $X \leftarrow Y$, $Y \leftarrow X$, or no causal influence between them, where $\leftarrow$ denotes causal direction. A key distinction between causal discovery and relation extraction in NLP is that causal discovery can reveal unknown causal relations, whereas relation extraction focuses on transforming relations in free text into structured relational tuples.

Although randomized controlled trials and A/B testing are the gold standard for causal discovery \cite{fisher:1935}, these experimental approaches are often impractical due to ethical or financial limitations. Thus, researchers turn to rely on statistical analysis of observational data to infer causal relations.

Statistical approaches to causal discovery can be broadly classified into: constraint-based methods, such as Peter and Clark (PC) \cite{spirtes2000causation} and inductive causation (IC) \cite{pearl2009causality}; score-based methods \cite{heckerman1995learning, chickering2002optimal, koivisto2004exact, 10.5555/2946645.2946677}; and functional methods \cite{shimizu06a, hyvarinen10a}. These methods employ statistical measures from observational data to construct causal graphs but have notable limitations. First, they require resource-intensive and extensive data collection. Second, theoretically, they cannot precisely identify ground-truth causal graphs but instead yield an equivalence class of true causal graphs \cite{spirtes2000causation, pearl2009causality}. 

Furthermore, many statistical approaches, such as PC and Greedy Equivalence Search (GES), operate under assumptions. \emph{Causal sufficiency} assumption posits that all variables are observed and included, neglecting the potential unobserved variables \cite{neal2020causalitybook}. Some algorithms, such as Tetrad condition-based \cite{JMLR:v7:silva06a,10.1145/2939672.2939838} and high-order moments-based approaches \cite{adams2021identification, chen2022identification} focus on only uncover specific types of unobserved variables, such as latent confounders (i.e., common causes). However, our work aims to identify more general unobserved variables, including confounders, mediators, causes, or effects of observed variables. \emph{Acyclicity} assumption states that causal graphs contain no cycles, which allows causal discovery to align with Bayesian network and simplifies mathematical challenges. However, this assumption often contradicts real-world phenomena. Many causal graphs are known to contain feedback loops, such as the poverty cycle:  poverty $\rightarrow$ limited access to education $\rightarrow$ low-paying jobs $\rightarrow$ poverty, \cite{banerjee2012poor, de2010breaking} and the predator-prey cycle: increase in predator population $\rightarrow$ decrease in prey population $\rightarrow$ decrease in predator population \cite{schmitz2017predator, 10.1093/oso/9780195131543.003.0028}. In contrast to prior work, our causal discovery framework allows for the inclusion of unobserved variables and permits cycles within causal graphs to align with real-world scenarios.

The advent of LLMs provides new opportunities to address causal discovery ~\cite{kıcıman2023causal, ze2023causal, long2022can}. These approaches require LLMs to determine the causal relation between a given pair of variable names. However, the reliability of such methods is under scrutiny. \citet{ze2023causal} argue that LLMs may function as \textit{"causal parrots"}, which depend on \textit{memorization} to recall the causal relations present in their training data rather than infer causal relations. This raises concerns about LLMs' \textit{generalization} to identify causal relations that are rare or absent in pre-training data. \citet{feng2024pretrainingcorporalargelanguage} presents empirical evidence that suggests while LLMs excel at reproducing frequent causal relations in pre-training data, they struggle to uncover novel causal relations. 

In contrast to approaches that directly employ LLMs for causal discovery, \citet{liu2024discoveryhiddenworldlarge} utilize LLMs to extract variables and their values from collected documents, then apply statistical methods to uncover causal relations among these variables. Our work diverges from this approach by only taking a set of initial variables as input and employing an automated process to collect relevant documents. After variable value extraction, we implement a hybrid causal discovery approach, which integrate both statistical and LLM-based methods. Furthermore, our framework is capable of identifying new variables that exhibit causal relations with the initial set, thereby enabling an iterative process of data collection and causal discovery on an expanded variables set. This iterative method allows for a comprehensive exploration of the causal relations surrounding the initial variables.

\section{Methodology}
\label{sec:methodology}
We introduce a real-time causal discovery framework, \ourmethod. Our method differs from prior causal discovery algorithms in three key aspects. First, \ourmethod does not rely on pre-existing observational data; instead, it automatically collects and extracts observational data related to the initial variables. Second, our hybrid causal discovery component can utilize known causal relations and uncover novel causal relations. Third, our approach relaxes the \emph{acyclicity} and \emph{causal sufficiency} assumptions.

\subsection{Problem Definition}
Given a set of initial variables, $\sZ=\{ \rz_1, \rz_2, ..., \rz_N \}$, where each $\rz_i$ represents one variable, the goal of real-time causal discovery is to automatically collect relevant unstructured data $\sD$ and extract variable values to form structured data $\sX$, which enables the discovery of causal relations through unstructured and structured data. After identifying causal relations among initial variables, the process involves identifying new variables causally related to the initial variables, resulting in an expanded set of variables $\sZ_{m}$. The final output is an expanded causal graph $\gG = (\sZ_{m}, \sR)$, where $\sR=\{\vr_1, ..., \vr_l\}$ represents the set of causal relations.

\subsection{Data Collection and Value Extraction}
\label{sec:data_collect_extract}

The first step of \ourmethod comprises two main steps: collection of relevant documents and extraction of variable values. The detailed procedure is outlined in \Algref{alg:data_collect_extract} in Appendix~\ref{apx:algorithms}.

\noindent  \textbf{Retrieval of Relevant Documents}
We retrieve relevant documents using the Google API \footnote{\url{https://developers.google.com/custom-search/docs/overview}}. To maximize the relevance to initial variables, we create search queries using a stepwise removal approach: 1) Begin with queries containing all variable names (\eg "smoking" AND "cancer" AND "pollution"). 2) Progressively remove one variable (\eg "smoking" AND "cancer"). 3) Stop with single-variable queries (\eg "smoking"). We also use synonyms of variables to enhance coverage. We select the top-k retrieved documents for each query. To ensure relevance to most variables, k is higher for queries containing more variables. The retrieval process continues until the total number of collected documents reaches a predefined threshold. The resulting document set is denoted as $\sD = \{d_1, .., d_T\}$, where $d_i$ represents one document.

\noindent  \textbf{Extraction of Variable Values}
We use LLMs to extract variable values from collected documents $\sD$. Given an LLM $\mM$, we design a prompt $l$ including a document $d_i$ and a description of one variable $\rz_j$. The variable description includes its name and the meaning of its values. We guide the LLM to generate responses following multiple thinking steps, simulating human expert reasoning, and provide the final answer in a specific format \cite{lin2024the}. This generation process can be denoted as $o_{ij}=\mM(l(d_i,\rz_j))$, where $o_{ij}$ is LLM's response regarding the value of variable $\rz_j$ in document $d_i$. We then extract the value $v_{ij}$ from response $o_{ij}$. By iterating through all variables and documents, we construct a structured data $\sX$ where each column represents a variable and each row represents a document. %The prompt template for value extraction is presented in Table~\ref{tab:prompt_value_extraction} in Appendix~\ref{apx:prompts}.

\subsection{Hybrid Causal Discovery}
\label{sec:hybrid_causal_discovery}
We employ a hybrid causal discovery approach, leveraging both statistical methods and LLM-based relation extraction techniques. The detailed process of our hybrid causal discovery method is outlined in \Algref{alg:hybrid_causal_discovery} in Appendix~\ref{apx:algorithms}.

\noindent \textbf{Statistical Causal Discovery}
For structured data $\sX$, we employ statistical causal discovery algorithms including PC \cite{spirtes2000causation}, GES \cite{10.1162/153244303321897717}, and NOTEARS \cite{10.5555/3327546.3327618}. For instance, the PC algorithm performs conditional independence tests between variable pairs, progressively expanding the conditioning sets to determine the presence of causal relations. These algorithms process structured data $\sX$ to produce a causal graph $\hat{\gG_{s}}$ as the output.

\noindent \textbf{LLM-based Causal Relation Extraction}
We introduce a novel causal relation extraction method inspired by causal relation verification \cite{si-etal-2024-checkwhy, wadden-etal-2022-scifact}. We treat each potential causal relation as a claim (\eg "smoking causes lung cancer") and find documents containing both the cause and effect terms (\eg "smoking" AND "lung cancer"). To ensure the trustworthiness of retrieved documents, we restrict the search domain to reputable academic repositories \footnote{Our search is limited to the following academic website domains: jstor.org, springer.com, ieee.org, ncbi.nlm.nih.gov, sciencedirect.com, scholar.google.com, arxiv.org.}. We then employ LLMs to assess whether each document supports or refutes or not relates with the causal relation using a carefully designed prompt. If a majority of documents support the causal relation, we incorporate it into a causal graph $\hat{\gG_{v}}$. Otherwise, it is excluded. 

\noindent  \textbf{Graph Merging}
The two branches of our hybrid method produce two causal graphs: $\hat{\gG_{s}}$ from statistical methods and $\hat{\gG_{v}}$ from the LLM-based approach. To merge them into the final causal graph $\hat{\gG}$, we post-process the causal graph $\hat{\gG_{s}}$ by adding high-confidence causal relations from $\hat{\gG_{v}}$ and removing those strongly refuted by the verification process. This merging strategy is employed for two reasons: (1) the structured data $\sX$ from the value extraction phase might contain noise; (2) causal relations that are widely supported or refuted by trustworthy documents can be treated as known knowledge.

\subsection{Missing Variable Proposal}
This step aims to identify missing variables not included in the initial set but potentially causally related to them, and append these to $\sZ_{m}$, as outlined in \Algref{alg:missing_variable_proposal} in Appendix~\ref{apx:algorithms}.

\noindent \textbf{Variable Abstraction}
We first use LLMs to abstract missing variables from the retrieved documents $\sD$. For each document, LLMs are instructed to analyze the content of each document, identify variables that could influence or be influenced by the initial variables, and then provide the most possible variable in a specified format. %The prompt is provided in Table~\ref{tab:prompt_variable_abstract} in Appendix~\ref{apx:prompts}.

\noindent \textbf{Variable Selection}
To select the most promising variables from all abstracted variables, we employ a dual approach combining causal relation verification and statistical measures.
\textit{Causal Relation Verification}: Using the method described in Section~\ref{sec:hybrid_causal_discovery}, we verify whether each new variable has a confirmed causal relation with any initial variable. Variables supported by the majority of documents are added to $\sZ_{m}$.
\textit{Statistical Measure}: We compute the Pointwise Mutual Information (PMI) between each new variable and the initial variables to quantify their dependence, with higher PMI scores indicating stronger potential causal association. The PMI between two variables $(\rz_i, \rz_j)$ is defined as:
\begin{equation}
\label{equ:pmi}
\begin{split}
    PMI(\rz_i, \rz_j) &= \log \frac{p(\rz_i, \rz_j)}{p(\rz_i)p(\rz_j)} = \log \frac{\frac{o(\rz_i, \rz_j)}{C}}{\frac{o(\rz_i)}{C}\frac{o(\rz_j)}{C}} \\
    &= \log \frac{o(\rz_i, \rz_j)}{o(\rz_i)o(\rz_j)} + \cancel{\log C}
\end{split}
\end{equation}
where $o(\rz_i, \rz_j)$ is the count of documents where $(\rz_i, \rz_j)$ co-occur, $o(\rz_i)$ is the count where $\rz_i$ appears, and $C$ is the total number of retrievable documents. Since $C$ is constant, $\log C$ is ignored. These counts are obtained by the Google Search API. We compute the PMI score of each abstracted variable with the initial variables and append the top $k$ variables with the highest aggregate PMI scores to $\sZ_{m}$.

With the expanded variables $\sZ_{m}$, we can iterate the data collection, value extraction, and causal discovery processes to generate an expanded causal graph $\gG = (\sZ_{m}, \sR)$ that incorporates these missing variables and new causal relations.

\section{Experiments}
\subsection{Evaluation of the IRIS Framework}
\label{sec:eval_expanded_causal_graphs}

% We evaluate the complete pipeline of \ourmethod on expanded causal graphs.

% % \subsubsection{Experimental Setup}
\noindent \textbf{Datasets.} The initial variables are from: Cancer \cite{10.5555/1941985}, Respiratory Disease, Diabetes, Obesity \cite{long2022can}, Alzheimer’s Disease Neuroimaging Initiative (ADNI) \cite{shen2020challenges}, and Insurance \cite{binder1997adaptive}. For more details, see Appendix~\ref{apx:ground_truth_causal_graphs}.

\noindent \textbf{Our Method and Baselines.} We employ GPT-4o due to its superior performance across all components of \ourmethod (see Sections~\ref{sec:eval_value_extraction}, \ref{sec:eval_causal_discovery}, and \ref{sec:eval_missing_variable_proposal}). All prompts in \ourmethod are designed using the Chain-of-Thought (CoT) \cite{wei2022chain} strategy and incorporate retrieved documents. Detailed prompt engineering for \ourmethod and baselines is provided in Appendix~\ref{apx:prompts}. For the statistical causal discovery algorithms, we utilize the Greedy Equivalence Search (GES) algorithm because it achieves the highest average F1 score and Normalized Hamming Distance (NHD) ratio across all datasets, as demonstrated in Section~\ref{sec:eval_causal_discovery}. 

We consider the following baselines: \textit{0-shot} relies solely on a zero-shot prompt. \textit{CoT} enhances prompts through a step-by-step reasoning process, mimicking human thought patterns. \textit{Retrieval-Augmented Generation (RAG)} \cite{lewis2020retrieval} incorporates retrieved documents into CoT prompt. Both baselines and human annotation determine causal relations among expanded variables from our missing variable proposal component.

\begin{table}[t]
\centering
\resizebox{\linewidth}{!}{%
\begin{tabular}{lccccc}
\hline
\rowcolor[HTML]{C0C0C0} 
Method     & P    & R    & \textbf{F1↑}  & Predict Edge & \textbf{NHD Ratio↓} \\ \hline
\rowcolor[HTML]{EFEFEF} 
\multicolumn{6}{c}{\cellcolor[HTML]{EFEFEF}Cancer}                            \\
0-shot     & 0.64 & 0.32 & 0.43          & 14           & 0.57                \\
CoT        & 0.67 & 0.38 & 0.48          & 18           & 0.54                \\
RAG        & 0.70 & 0.44 & 0.54          & 17           & 0.49                \\
\ourmethod & 0.89 & 0.57 & \textbf{0.70} & 18           & \textbf{0.30}       \\ \hline
\rowcolor[HTML]{EFEFEF} 
\multicolumn{6}{c}{\cellcolor[HTML]{EFEFEF}Respiratory Disease}                       \\
0-shot     & 0.67 & 0.36 & 0.47          & 12           & 0.53                \\
CoT        & 0.64 & 0.4  & 0.49          & 12           & 0.51                \\
RAG        & 0.64 & 0.45 & 0.53          & 16           & 0.47                \\
\ourmethod & 0.67 & 0.55 & \textbf{0.60} & 18           & \textbf{0.40}       \\ \hline
\rowcolor[HTML]{EFEFEF} 
\multicolumn{6}{c}{\cellcolor[HTML]{EFEFEF}Diabetes}                          \\
0-shot     & 0.70 & 0.46 & 0.56          & 17           & 0.45                \\
CoT        & 0.66 & 0.48 & 0.55          & 16           & 0.46                \\
RAG        & 0.73 & 0.47 & 0.57          & 16           & 0.43                \\
\ourmethod & 0.76 & 0.50 & \textbf{0.60} & 17           & \textbf{0.39}       \\ \hline
\rowcolor[HTML]{EFEFEF} 
\multicolumn{6}{c}{\cellcolor[HTML]{EFEFEF}Obesity}                           \\
0-shot     & 0.57 & 0.33 & 0.42          & 14           & 0.58                \\
CoT        & 0.59 & 0.38 & 0.46          & 25           & 0.54                \\
RAG        & 0.62 & 0.45 & 0.52          & 19           & 0.49                \\
\ourmethod & 0.67 & 0.58 & \textbf{0.62} & 21           & \textbf{0.38}       \\ \hline
\rowcolor[HTML]{EFEFEF} 
\multicolumn{6}{c}{\cellcolor[HTML]{EFEFEF}ADNI}                              \\
0-shot     & 0.47 & 0.29 & 0.36          & 17           & 0.64                \\
CoT        & 0.46 & 0.31 & 0.37          & 21           & 0.62                \\
RAG        & 0.50 & 0.34 & 0.40          & 19           & 0.60                \\
\ourmethod & 0.50 & 0.36 & \textbf{0.42} & 20           & \textbf{0.58}       \\ \hline
\rowcolor[HTML]{EFEFEF} 
\multicolumn{6}{c}{\cellcolor[HTML]{EFEFEF}Insurance}                         \\
0-shot     & 0.35 & 0.38 & 0.36          & 69           & 0.65                \\
CoT        & 0.41 & 0.38 & 0.39          & 65           & 0.61                \\
RAG        & 0.44 & 0.40 & 0.42          & 67           & 0.57                \\
\ourmethod & 0.61 & 0.46 & \textbf{0.53} & 49           & \textbf{0.47}       \\ \hline
\end{tabular}%
}
\caption{Evaluation results of the complete framework. }
\label{tab:result_expanded_causal_graphs}
% \vspace{-10pt}
\end{table}

\noindent \textbf{Evaluation.} We hire three domain experts to independently annotate ground-truth expanded causal graphs. Edges are included if at least two annotators agree. Inter-annotator agreement is high, with a Krippendorff’s alpha of 0.88 \cite{Krippendorff2011ComputingKA}. The detailed annotation instruction is in Table~\ref{tab:causal_relation_annotation_instruction} in Appendix~\ref{apx:causal_relation_annotation_task}. The expanded causal graphs are illustrated in Figures \ref{fig:expanded_causal_graphs_cancer} - \ref{fig:expanded_causal_graphs_adni} in Appendix~\ref{apx:expanded_causal_graphs}.  Following \citet{kıcıman2023causal, feng2024pretrainingcorporalargelanguage}, we evaluate the results of causal discovery using precision, recall, F1 score, and the Ratio of Normalized Hamming Distance (NHD) to baseline NHD. The ratio is defined as \( \text{ratio} = \frac{\text{NHD}}{\text{baseline NHD}} \), where the \( \text{baseline NHD} \) is derived from the worst-performing causal graph that has the same number of edges as the predicted graph. A lower ratio signifies a more accurate predicted causal graph. All results are averaged over three independent runs per causal graph.

% \subsubsection{Experimental Results and Analysis}
\noindent \textbf{Experimental Results and Analysis.}
Table~\ref{tab:result_expanded_causal_graphs} demonstrates that \ourmethod consistently outperforms all baselines across all datasets, achieving the highest F1 scores and lowest NHD ratios. A paired t-test \cite{Ross2017} confirms that the performance differences between \ourmethod and the baselines (in both F1 and NHD ratio) are statistically significant (\textit{p}-value $\leq 0.05$). For both the baselines and \ourmethod, the variance across all metrics is below 0.05, likely due to the consistency of the retrieved documents and the stability of GPT-4o’s responses. In terms of precision and recall, while some baselines (e.g., RAG in ADNI) achieve comparable precision to \ourmethod, none match its recall. This highlights \ourmethod's ability to uncover a greater number of true causal relations through its hybrid causal discovery approach. Among the datasets, ADNI exhibits the lowest overall performance for both methods, likely due to the inherent complexity of Alzheimer's disease causal relations. Meanwhile, the Insurance dataset, which contains the most complex causal graph (expanding from 27 initial variables to 35 variables and 67 edges), showcases the scalability of \ourmethod. Among the baselines, RAG performs better than others, underscoring the effectiveness of integrating retrieved documents with reasoning steps for causal discovery.

\subsection{Evaluation of Value Extraction}
\label{sec:eval_value_extraction}

% \subsubsection{Experimental Setup}
\noindent \textbf{Datasets.}
We evaluate the value extraction component of our method using two table-to-text datasets: AppleGastronome and Neuropathic \cite{liu2024discoveryhiddenworldlarge}. These datasets are particularly suitable for our task as they provide tabular data where columns represent variables and rows represent samples. Each row is associated with a corresponding textual description. The datasets are structured as follows: AppleGastronome contains 7 variables and 100 samples. Variable values are -1, 0, or 1. Neuropathic contains 7 variables and 100 samples. Variable values are 0 or 1.

\noindent \textbf{Our Method and Baselines.}
We utilize state-of-the-art LLMs for our method:
Llama-3.1-8b-Instruct \cite{Llama31}, GPT-3.5-turbo \cite{chatgpt}, GPT-4o \cite{gpt4o}. Additionally, we compare our method with COAT, which also utilizes an LLM to extract values of variables from documents \cite{liu2024discoveryhiddenworldlarge}. To ensure a fair comparison, we use GPT-4o in both our method and the COAT implementation.

\noindent \textbf{Metrics.}
Given that variable values are categorical, we frame the value extraction task as a classification problem, predicting the value of a variable in a given document. Therefore, we employ standard classification metrics: precision, recall, and F1.

\begin{table}[t]
\centering
% \resizebox{\textwidth}{!}{%
\begin{tabular}{lccc}
\hline
\rowcolor[HTML]{C0C0C0} 
\multicolumn{1}{c}{\cellcolor[HTML]{C0C0C0}Method} & P             & R             & F1            \\ \hline
\rowcolor[HTML]{EFEFEF} 
\multicolumn{4}{c}{\cellcolor[HTML]{EFEFEF}AppleGastronome}                                        \\
COAT (GPT-4o)                                      & 0.74          & 0.76          & 0.75          \\
\ourmethod (Llama)                                 & 0.71          & 0.72          & 0.71          \\
\ourmethod (GPT-3.5)                               & 0.75          & 0.77          & 0.76          \\
\ourmethod (GPT-4o)                                & \textbf{0.79} & \textbf{0.82} & \textbf{0.79} \\ \hline
\rowcolor[HTML]{EFEFEF} 
\multicolumn{4}{c}{\cellcolor[HTML]{EFEFEF}Neuropathic}                                            \\
COAT (GPT-4o)                                      & 0.72          & 0.80          & 0.79          \\
\ourmethod (Llama)                                 & \textbf{0.76} & 0.82          & 0.79          \\
\ourmethod (GPT-3.5)                               & 0.71          & 0.89          & 0.79          \\
\ourmethod (GPT-4o)                                & 0.73          & \textbf{1.0}  & \textbf{0.84} \\ \hline
\end{tabular}%
% }
\caption{Evaluation results of value extraction. Llama represents Llama-3.1-8b-instruct.}
\label{tab:result_value_extraction}
% \vspace{-15pt}
\end{table}

% \subsubsection{Experimental Results and Analysis}
\noindent \textbf{Experimental Results and Analysis.}
Table~\ref{tab:result_value_extraction} presents the evaluation results of the value extraction component on the AppleGastronome and Neuropathic datasets. Our method's superior performance with GPT-4o, compared to COAT using the same LLM, indicates that our approach is more effective than COAT under identical LLM. In both datasets, we observe a consistent trend of improvement from Llama-3.1-8b-Instruct to GPT-3.5, and further to GPT-4o when using our method. This progression aligns with the general understanding that more advanced LLMs tend to perform better on complex tasks. Overall, the models perform better on the Neuropathic dataset compared to AppleGastronome. This could be attributed to the simpler binary values of the Neuropathic dataset (values 0 or 1) compared to the ternary values in AppleGastronome (-1, 0, 1). The additional complexity in AppleGastronome might introduce more opportunities for misclassification. The high performance of GPT-4o suggests that it could be highly effective for value extraction in our framework.

\subsection{Evaluation of Causal Discovery}
\label{sec:eval_causal_discovery}
% In this section, we evaluate the causal discovery capabilities of our method.

\begin{figure*}[ht]
    \centering
    \includegraphics[width=\textwidth]{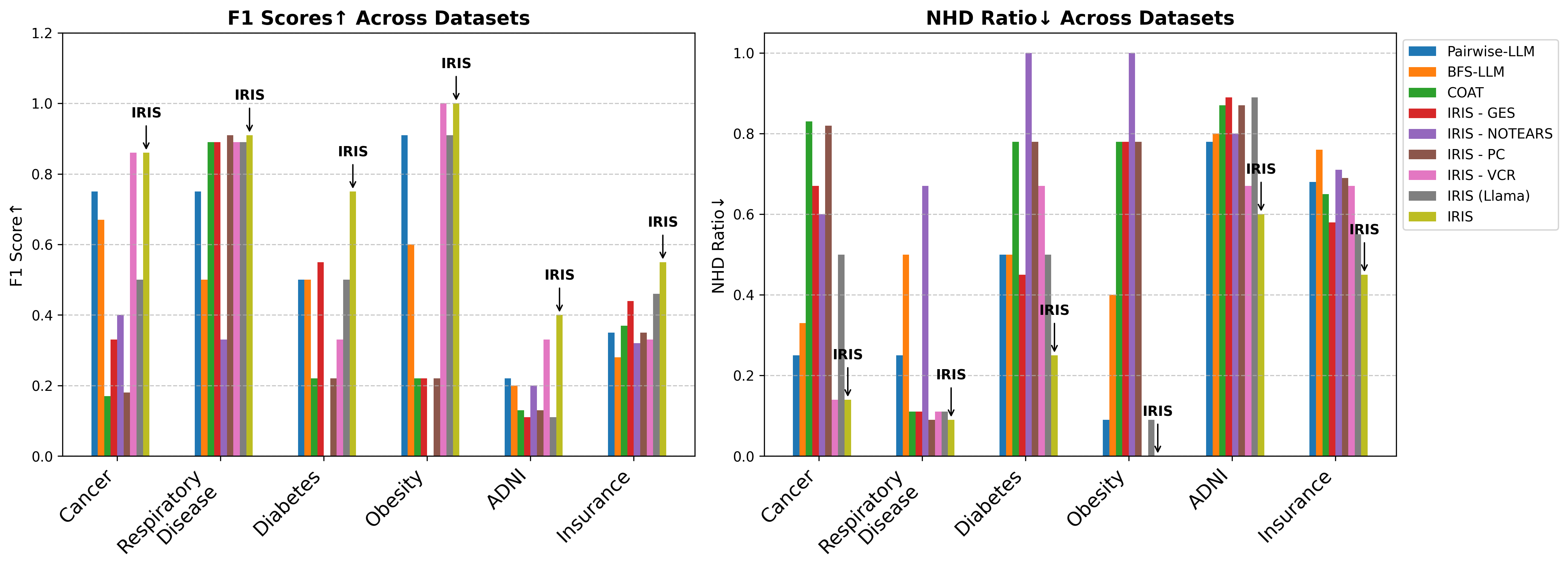}
    \caption{Evaluation results of causal discovery component on five datasets. A higher F1 score indicates better performance, while a lower NHD ratio reflects better performance. VCR refers to verified causal relations that are extracted from relevant academic documents and validated by LLMs. "Llama" refers to the use of the Llama-3.1-8b-instruct model as a substitute for GPT-4o in our method.}
    \label{fig:result_causal_discovery_all}
% \vspace{-15pt}
\end{figure*}

% \subsubsection{Experimental Setup}
\noindent \textbf{Datasets.}
We evaluate our hybrid causal discovery component on: Cancer \cite{10.5555/1941985}, Respiratory Disease, Diabetes, Obesity \cite{long2022can}, Alzheimer’s Disease Neuroimaging Initiative (ADNI) \cite{shen2020challenges}, and Insurance \cite{binder1997adaptive}.  These causal graphs are annotated by domain experts. The ground-truth causal graphs are presented in \Figref{fig:ground_truth_causal_graphs} in Appendix~\ref{apx:ground_truth_causal_graphs}.

\noindent \textbf{Our Method and Baselines.}
In our hybrid causal discovery, for statistical algorithms, we utilize PC \cite{spirtes2000causation}, GES \cite{10.1162/153244303321897717}, and NOTEARS \cite{10.5555/3327546.3327618}. Among the three statistical methods, we select the one that demonstrates the best performance for hybrid causal discovery. Based on the value extraction results (see Table~\ref{tab:result_value_extraction}), we use GPT-4o, which demonstrated the best performance, as the LLM for both our method and the baseline approaches. To illustrate how different LLMs affect the performance of our method, we employ the Llama-3.1-8b-instruct model as a counterpart. We compare our method against several baselines: 1) Pairwise-LLM \cite{feng2024pretrainingcorporalargelanguage} calls LLMs for each pair of variables to determine causal relations. 2) BFS-LLM \cite{jiralerspong2024efficientcausalgraphdiscovery} employs a breadth-first search with LLMs to determine causal relations. 3) COAT extracts values using LLMs and discovers causal relations with the PC algorithm \cite{liu2024discoveryhiddenworldlarge}.

\noindent \textbf{Metrics.} We evaluate predicted causal graphs using precision, recall, F1, and NHD ratio as detailed in Section~\ref{sec:eval_expanded_causal_graphs}. 

%Following \citet{kıcıman2023causal, feng2024pretrainingcorporalargelanguage}, we evaluate the results of causal discovery using precision, recall, F1 score, and the Ratio of Normalized Hamming Distance (NHD) to baseline NHD. The ratio is defined as \( \text{ratio} = \frac{\text{NHD}}{\text{baseline NHD}} \), where the \( \text{baseline NHD} \) is derived from the worst-performing causal graph that has the same number of edges as the predicted graph. A lower ratio signifies a more accurate predicted causal graph.

% \input{tables/result_causal_discovery_cancer}
% \input{tables/result_causal_discovery_respiratory_disease}
% \input{tables/result_causal_discovery_diabetes}
% \input{tables/result_causal_discovery_obesity}
% \input{tables/result_causal_discovery_adni}

% \subsubsection{Experimental Results and Analysis}
\noindent \textbf{Experimental Results and Analysis.}
The evaluation results of the causal discovery component across datasets are presented in Figure~\ref{fig:result_causal_discovery_all}. 
More detailed are presented in Table~\ref{tab:result_causal_discovery_cancer} - ~\ref{tab:result_causal_discovery_insurance} in Appendix~\ref{apx:eval_causal_discovery_component}. In these results, our hybrid method consistently outperforms baselines across all datasets. This highlights the effectiveness of combining statistical algorithms with LLM-based methods.

We observe that the performance of individual statistical algorithms (GES, NOTEARS, PC) varied across datasets. PC excels in Respiratory Disease and Obesity. GES demonstrates optimal performance on Diabetes and Obesity. NOTEARS performs best on Cancer and ADNI but struggles significantly with Diabetes and Obesity, achieving a 0 F1 score and a 1 NHD ratio. This variation highlights the importance of selecting statistical algorithms based on the characteristics of the observational data, which presents a compelling area for further research. From our experiments, both GES and PC exhibit strong performances; however, GES outperforms PC with a 0.09 higher average F1 score and a 0.09 lower average NHD ratio. Given these results, GES is recommended as the primary choice when the suitability of the algorithm is uncertain. When comparing the performance of Llama-3.1-8b-instruct and GPT-4o, GPT-4o consistently outperforms Llama-3.1-8b-instruct across all datasets, with a particularly significant gap observed in the ADNI dataset. We believe this discrepancy arises because ADNI involves specialized knowledge that is less commonly represented in the pre-training data of Llama-3.1-8b-instruct.

Pairwise-LLM and BFS-LLM show competitive performance on simpler datasets. They perform well on the Cancer and Respiratory Disease datasets. However, their performance degrades on more complex datasets like ADNI. This suggests that while LLMs have potential in causal discovery, they may struggle with more complex causal relations, possibly due to the lower occurrence of such domain-specific causal relations in their training data \cite{feng2024pretrainingcorporalargelanguage}. The COAT method yields results similar to \ourmethod - PC because both approaches extract values from documents and discover causal relations through the PC algorithm.

\subsection{Evaluation of Missing Variable Proposal}
\label{sec:eval_missing_variable_proposal}

% \subsubsection{Experimental Setup}
\textbf{Datasets.} Evaluating the missing variable proposal component presents a unique challenge: the ground-truth missing variables are inherently unknown in real-world scenarios. To address this, we simulate missing variables and assess our method's ability to identify them. We start with complete, ground-truth causal graphs and remove variables to create incomplete graphs. We employ the initial causal graphs from Cancer, Respiratory Disease, Diabetes, Obesity, ADNI, and Insurance. For each causal graph, we iteratively remove one variable at a time, creating multiple test cases per graph. We then apply our missing variable proposal component to these incomplete graphs, aiming to identify the removed variables.

\noindent \textbf{Our Method and Baselines.} For our method and the baseline, we use GPT-4o as the primary LLM. To assess the impact of different LLMs, we also replace GPT-4o with Llama-3.1-8b-instruct in our method. We compare with the following baselines: \textit{0-shot}, which generates new variables using a zero-shot prompt; \textit{CoT}, which enhances the prompt with reasoning steps; and \textit{RAG}, which proposes new variables based on retrieved documents, similar to our method but relying solely on prompting to select the final variables. Prompts of baselines are provided in Appendix~\ref{apx:prompts}.

\noindent \textbf{Metrics.} We evaluate the performance using a \textit{success rate} metric, calculated as follows: 1) For each incomplete causal graph, we check if our method successfully proposes the removed variable in its final set of proposed variables $\sZ_{m}$. 2) We count a "success" for each correctly proposed variable. 3) The success rate is computed as: Success Rate = Number of Successes / Total Number of Incomplete Graphs. For instance, in a causal graph with five variables, we create five different incomplete graphs by removing each variable. If our method correctly proposes the removed variable in three of these five graphs, the success rate would be 0.6 (=3/5). For the statistical approach, we select the top-5 variables based on their PMI scores.

\begin{table}[t]
\centering
\resizebox{\linewidth}{!}{%
\begin{tabular}{lcccccc}
\hline
\rowcolor[HTML]{C0C0C0} 
Method                     & Cancer        & \begin{tabular}[c]{@{}c@{}}Resp. \\ Disease\end{tabular} & Diabetes      & Obesity       & ADNI          & Insurance     \\ \hline
0-shot                     & 0.40          & 0.25                                                     & 0.50          & 0.25          & 0.25          & 0.22          \\
CoT                        & 0.40          & 0.50                                                     & 0.50          & 0.75          & 0.25          & 0.30          \\
RAG                        & 0.60          & 0.75                                                     & 0.75          & 0.75          & 0.38          & 0.41          \\
MVP                        & \textbf{0.80} & \textbf{0.75}                                            & \textbf{1.00} & \textbf{1.00} & \textbf{0.50} & \textbf{0.59} \\
~~- VCR                    & 0.60          & 0.75                                                     & 0.50          & 0.75          & 0.25          & 0.48          \\
~~- Stats                  & 0.60          & 0.75                                                     & 0.75          & 1.00          & 0.38          & 0.52          \\
~~ $\leftrightarrow$ Llama & 0.40          & 0.50                                                     & 0.25          & 0.50          & 0.13          & 0.45          \\ \hline
\end{tabular}%
}
\caption{Success rate of the missing variable proposal (MVP) component. -VCR omits verified causal relation extraction; -Stats omits statistical approaches; $\leftrightarrow$ Llama uses Llama-3.1-8b-instruct instead of GPT-4o.}
\label{tab:results_missing_variable_proposal}
% \vspace{-10pt}
\end{table}

% \subsubsection{Experimental Results and Analysis}
\noindent \textbf{Experimental Results and Analysis.}
The evaluation results of our Missing Variable Proposal (MVP) component are presented in Table~\ref{tab:results_missing_variable_proposal}. The MVP method consistently outperforms other baselines and ablation variants across all datasets. This demonstrates the effectiveness of combining verified causal relation extraction (VCR) with statistical approach (Stats) in identifying missing variables. Ablation studies indicate that both VCR and statistical approaches play a crucial role in enhancing the success rate of the MVP. The performance gap between MVP and MVP $\leftrightarrow$ Llama indicates the superior capability of GPT-4o in understanding and reasoning about causal relations. All baselines consistently underperform compared to our MVP, indicating that relying solely on the textual knowledge from documents and LLMs is not enough for proposing missing variables.

\section{Conclusion}
\label{sec:conclusion}
In this paper, we introduce IRIS, a novel framework that addresses several longstanding challenges in causal discovery. By integrating automated data collection, hybrid causal discovery methods, and a variable proposal components, IRIS significantly advances our ability to uncover causal relations in real-world scenarios. Our approach not only reduces the reliance on extensive manual data collection but also leverages existing knowledge in order to facilitate the discovery of novel causal relations with novel variables. Our experimental results show that IRIS consistently outperforms competitive baselines. Future work could aim to enhancing the scalability of IRIS for larger and more complex causal graphs by integrating causal relations extracted from texts with the ones identified through statistical algorithms.

\section*{Limitations}
Our approach to uncovering causal relations using retrieved documents and LLMs has certain limitations. A primary challenge is the potential bias inherent in both the data and the LLMs. Retrieved documents may contain sampling biases, inaccuracies, or incomplete coverage of causal relations. Likewise, LLMs may inherit biases from their pre-training data or face limitations in generalization, potentially affecting their interpretation of causal relationships. To mitigate these issues, we retrieve documents from reliable academic websites, and leverage state-of-the-art LLMs like GPT-4o.

Another limitation is that the number of queries to the LLM grows quadratically with the number of variables. On average, our method takes approximately 15 hours to run, which is about three times slower than the zero-shot baseline. However, all LLM-related processes can be parallelized. For instance, in causal relation extraction, each causal relation can be independently queried in parallel to determine whether the relation holds. This parallel processing significantly mitigates the computational overhead and ensures that the framework remains scalable even as the number of variables increases. 

Finally, the energy consumption of LLM inference presents an environmental challenge. While optimizing efficiency in LLM inference is an important research direction, it is beyond the scope of this work.

\section*{Ethics Statement}
% \subsection{Ethics Statement}
We acknowledge the importance of ACL Code of Ethics and agree with it. We ensure that our study is compatible with the provided code.

Our work involves uncovering causal relations using retrieved documents and LLMs, and we acknowledge the ethical considerations associated with this approach. The potential biases inherent in both the retrieved data and the LLMs pose a significant challenge. To mitigate these risks, we prioritize retrieving data from credible sources, such as academic publications and verified websites, to ensure the reliability of the input data. Additionally, we employ state-of-the-art LLMs, like GPT-4, which are designed to provide high-quality and robust outputs. However, we recognize that no system is entirely free from bias, and users of this framework should exercise caution in interpreting its results.

The evaluation of our method involves hiring human experts to annotate causal graphs. We have ensured that the annotation process adheres to ethical guidelines, including providing fair compensation for their contributions. Rigorous measures have been taken to thoroughly anonymize the causal graphs, which do not contain any personally identifiable information or sensitive data related to the contributors. The causal graphs were compiled with contributions from PhD students, which may inherently introduce biases influenced by their demographic backgrounds. We advise researchers utilizing this dataset to carefully account for these potential biases, particularly in studies related to AI fairness, bias, and safety.

% Bibliography entries for the entire Anthology, followed by custom entries
%\bibliography{anthology,custom}
% Custom bibliography entries only
\bibliography{custom}

\begin{thebibliography}{65}
\providecommand{\natexlab}[1]{#1}

\bibitem[{Abrams(2001)}]{10.1093/oso/9780195131543.003.0028}
Peter~A Abrams. 2001.
\newblock \href {https://doi.org/10.1093/oso/9780195131543.003.0028} {{Predator-Prey Interactions}}.
\newblock In \emph{{Evolutionary Ecology: Concepts and Case Studies }}. Oxford University Press.

\bibitem[{Adams et~al.(2021)Adams, Hansen, and Zhang}]{adams2021identification}
Jeffrey Adams, Niels~Richard Hansen, and Kun Zhang. 2021.
\newblock \href {https://openreview.net/forum?id=yCA2i3bGbfC} {Identification of partially observed linear causal models: Graphical conditions for the non-gaussian and heterogeneous cases}.
\newblock In \emph{Advances in Neural Information Processing Systems}.

\bibitem[{Asai et~al.(2024)Asai, Wu, Wang, Sil, and Hajishirzi}]{asai2024selfrag}
Akari Asai, Zeqiu Wu, Yizhong Wang, Avirup Sil, and Hannaneh Hajishirzi. 2024.
\newblock \href {https://openreview.net/forum?id=hSyW5go0v8} {Self-{RAG}: Learning to retrieve, generate, and critique through self-reflection}.
\newblock In \emph{The Twelfth International Conference on Learning Representations}.

\bibitem[{Balashankar et~al.(2019)Balashankar, Chakraborty, Fraiberger, and Subramanian}]{balashankar-etal-2019-identifying}
Ananth Balashankar, Sunandan Chakraborty, Samuel Fraiberger, and Lakshminarayanan Subramanian. 2019.
\newblock Identifying predictive causal factors from news streams.
\newblock In \emph{EMNLP:2019:1}, pages 2338--2348, Hong Kong, China. acl.

\bibitem[{Banerjee and Duflo(2012)}]{banerjee2012poor}
A.V. Banerjee and E.~Duflo. 2012.
\newblock \href {https://books.google.com.au/books?id=2dlnBoX4licC} {\emph{Poor Economics: A Radical Rethinking of the Way to Fight Global Poverty}}.
\newblock PublicAffairs.

\bibitem[{Bazaga et~al.(2024)Bazaga, Lio, and Micklem}]{bazaga2024unsupervised}
Adri{\'a}n Bazaga, Pietro Lio, and Gos Micklem. 2024.
\newblock \href {https://openreview.net/forum?id=1mjsP8RYAw} {Unsupervised pretraining for fact verification by language model distillation}.
\newblock In \emph{The Twelfth International Conference on Learning Representations}.

\bibitem[{Bekoulis et~al.(2021)Bekoulis, Papagiannopoulou, and Deligiannis}]{bekoulis2021review}
Giannis Bekoulis, Christina Papagiannopoulou, and Nikos Deligiannis. 2021.
\newblock A review on fact extraction and verification.
\newblock \emph{ACM Computing Surveys (CSUR)}, 55(1):1--35.

\bibitem[{Binder et~al.(1997)Binder, Koller, Russell, and Kanazawa}]{binder1997adaptive}
John Binder, Daphne Koller, Stuart Russell, and Keiji Kanazawa. 1997.
\newblock Adaptive probabilistic networks with hidden variables.
\newblock \emph{Machine Learning}, 29:213--244.

\bibitem[{Bui et~al.(2010)Bui, Nuall{\'a}in, Boucher, and Sloot}]{bui2010extracting}
Quoc-Chinh Bui, Breannd{\'a}n~{\'O} Nuall{\'a}in, Charles~A Boucher, and Peter~MA Sloot. 2010.
\newblock \href {https://pubmed.ncbi.nlm.nih.gov/20178611/} {Extracting causal relations on hiv drug resistance from literature}.
\newblock \emph{BMC bioinformatics}, 11:1--11.

\bibitem[{Chang and Choi(2006)}]{CHANG2006662}
Du-Seong Chang and Key-Sun Choi. 2006.
\newblock \href {https://doi.org/10.1016/j.ipm.2005.04.004} {Incremental cue phrase learning and bootstrapping method for causality extraction using cue phrase and word pair probabilities}.
\newblock \emph{Information Processing and Management}, 42(3):662--678.

\bibitem[{Chen et~al.(2022)Chen, Xie, Qiao, Hao, Zhang, and Cai}]{chen2022identification}
Zhengming Chen, Feng Xie, Jie Qiao, Zhifeng Hao, Kun Zhang, and Ruichu Cai. 2022.
\newblock Identification of linear latent variable model with arbitrary distribution.
\newblock In \emph{Proceedings of the AAAI Conference on Artificial Intelligence}, volume~36, pages 6350--6357.

\bibitem[{Chickering(2002)}]{chickering2002optimal}
David~Maxwell Chickering. 2002.
\newblock Optimal structure identification with greedy search.
\newblock \emph{Journal of machine learning research}, 3(Nov):507--554.

\bibitem[{Chickering(2003)}]{10.1162/153244303321897717}
David~Maxwell Chickering. 2003.
\newblock \href {https://doi.org/10.1162/153244303321897717} {Optimal structure identification with greedy search}.
\newblock \emph{J. Mach. Learn. Res.}, 3(null):507–554.

\bibitem[{Dasgupta et~al.(2018)Dasgupta, Saha, Dey, and Naskar}]{dasgupta-etal-2018-automatic-extraction}
Tirthankar Dasgupta, Rupsa Saha, Lipika Dey, and Abir Naskar. 2018.
\newblock Automatic extraction of causal relations from text using linguistically informed deep neural networks.
\newblock In \emph{WS:2018:50}, pages 306--316, Melbourne, Australia. acl.

\bibitem[{De~Weiss and Sirkin(2010)}]{de2010breaking}
Susan~Pick De~Weiss and Jenna Sirkin. 2010.
\newblock \emph{Breaking the poverty cycle: The human basis for sustainable development}.
\newblock Oxford University Press.

\bibitem[{Feng et~al.(2023)Feng, Qu, and Haffari}]{feng-etal-2023-less}
Tao Feng, Lizhen Qu, and Gholamreza Haffari. 2023.
\newblock \href {https://doi.org/10.1162/tacl_a_00561} {Less is more: Mitigate spurious correlations for open-domain dialogue response generation models by causal discovery}.
\newblock \emph{Transactions of the Association for Computational Linguistics}, 11:511--530.

\bibitem[{Feng et~al.(2025)Feng, Qu, Kang, and Haffari}]{feng-etal-2025-causalscore}
Tao Feng, Lizhen Qu, Xiaoxi Kang, and Gholamreza Haffari. 2025.
\newblock {C}ausal{S}core: An automatic reference-free metric for assessing response relevance in open-domain dialogue systems.
\newblock In \emph{COLING:2025:main}, pages 2351--2369, Abu Dhabi, UAE. acl.

\bibitem[{Feng et~al.(2024{\natexlab{a}})Feng, Qu, Li, Zhan, Hua, and Haf}]{feng-etal-2024-imo}
Tao Feng, Lizhen Qu, Zhuang Li, Haolan Zhan, Yuncheng Hua, and Reza Haf. 2024{\natexlab{a}}.
\newblock \href {https://doi.org/10.18653/v1/2024.acl-long.144} {{IMO}: Greedy layer-wise sparse representation learning for out-of-distribution text classification with pre-trained models}.
\newblock In \emph{ACL:2024:long}, pages 2625--2639, Bangkok, Thailand. acl.

\bibitem[{Feng et~al.(2024{\natexlab{b}})Feng, Qu, Tandon, Li, Kang, and Haffari}]{feng2024pretrainingcorporalargelanguage}
Tao Feng, Lizhen Qu, Niket Tandon, Zhuang Li, Xiaoxi Kang, and Gholamreza Haffari. 2024{\natexlab{b}}.
\newblock \href {https://arxiv.org/abs/2407.19638} {From pre-training corpora to large language models: What factors influence llm performance in causal discovery tasks?}
\newblock \emph{Preprint}, arXiv:2407.19638.

\bibitem[{Fisher(1935)}]{fisher:1935}
R.~A. Fisher. 1935.
\newblock \emph{The Design of Experiments}.
\newblock Oliver and Boyd.

\bibitem[{Garcia(1997)}]{10.5555/645359.651023}
Daniela Garcia. 1997.
\newblock \href {https://dl.acm.org/doi/10.5555/645359.651023} {Coatis, an nlp system to locate expressions of actions connected by causality links}.
\newblock In \emph{Proceedings of the 10th European Workshop on Knowledge Acquisition, Modeling and Management}, EKAW '97, page 347–352, Berlin, Heidelberg. Springer-Verlag.

\bibitem[{Glymour et~al.(2019)Glymour, Zhang, and Spirtes}]{10.3389/fgene.2019.00524}
Clark Glymour, Kun Zhang, and Peter Spirtes. 2019.
\newblock \href {https://doi.org/10.3389/fgene.2019.00524} {Review of causal discovery methods based on graphical models}.
\newblock \emph{Frontiers in Genetics}, 10.

\bibitem[{Heckerman et~al.(1995)Heckerman, Geiger, and Chickering}]{heckerman1995learning}
David Heckerman, Dan Geiger, and David~M Chickering. 1995.
\newblock Learning bayesian networks: The combination of knowledge and statistical data.
\newblock \emph{Machine learning}, 20:197--243.

\bibitem[{Heindorf et~al.(2020)Heindorf, Scholten, Wachsmuth, Ngonga~Ngomo, and Potthast}]{10.1145/3340531.3412763}
Stefan Heindorf, Yan Scholten, Henning Wachsmuth, Axel-Cyrille Ngonga~Ngomo, and Martin Potthast. 2020.
\newblock \href {https://doi.org/10.1145/3340531.3412763} {Causenet: Towards a causality graph extracted from the web}.
\newblock In \emph{Proceedings of the 29th ACM International Conference on Information \& Knowledge Management}, CIKM '20, page 3023–3030, New York, NY, USA. Association for Computing Machinery.

\bibitem[{Hern{\'a}n et~al.(2004)Hern{\'a}n, Hern{\'a}ndez-D{\'\i}az, and Robins}]{hernan2004structural}
Miguel~A Hern{\'a}n, Sonia Hern{\'a}ndez-D{\'\i}az, and James~M Robins. 2004.
\newblock A structural approach to selection bias.
\newblock \emph{Epidemiology}, 15(5):615--625.

\bibitem[{Hyv{{\"a}}rinen et~al.(2010)Hyv{{\"a}}rinen, Zhang, Shimizu, and Hoyer}]{hyvarinen10a}
Aapo Hyv{{\"a}}rinen, Kun Zhang, Shohei Shimizu, and Patrik~O. Hoyer. 2010.
\newblock \href {http://jmlr.org/papers/v11/hyvarinen10a.html} {Estimation of a structural vector autoregression model using non-gaussianity}.
\newblock \emph{Journal of Machine Learning Research}, 11(56):1709--1731.

\bibitem[{Jiralerspong et~al.(2024)Jiralerspong, Chen, More, Shah, and Bengio}]{jiralerspong2024efficientcausalgraphdiscovery}
Thomas Jiralerspong, Xiaoyin Chen, Yash More, Vedant Shah, and Yoshua Bengio. 2024.
\newblock \href {https://arxiv.org/abs/2402.01207} {Efficient causal graph discovery using large language models}.
\newblock \emph{Preprint}, arXiv:2402.01207.

\bibitem[{Khoo et~al.(2000)Khoo, Chan, and Niu}]{khoo-etal-2000-extracting}
Christopher S.~G. Khoo, Syin Chan, and Yun Niu. 2000.
\newblock Extracting causal knowledge from a medical database using graphical patterns.
\newblock In \emph{ACL:2000:1}, pages 336--343, Hong Kong. acl.

\bibitem[{Kim et~al.(2024)Kim, Lee, Huang, Chan, Li, and Ji}]{kim2024llmsproducefaithfulexplanations}
Kyungha Kim, Sangyun Lee, Kung-Hsiang Huang, Hou~Pong Chan, Manling Li, and Heng Ji. 2024.
\newblock \href {https://arxiv.org/abs/2402.07401} {Can llms produce faithful explanations for fact-checking? towards faithful explainable fact-checking via multi-agent debate}.
\newblock \emph{Preprint}, arXiv:2402.07401.

\bibitem[{Koivisto and Sood(2004)}]{koivisto2004exact}
Mikko Koivisto and Kismat Sood. 2004.
\newblock Exact bayesian structure discovery in bayesian networks.
\newblock \emph{The Journal of Machine Learning Research}, 5:549--573.

\bibitem[{Korb and Nicholson(2010)}]{10.5555/1941985}
Kevin~B. Korb and Ann~E. Nicholson. 2010.
\newblock \emph{Bayesian Artificial Intelligence, Second Edition}, 2nd edition.
\newblock CRC Press, Inc., USA.

\bibitem[{Krippendorff(2011)}]{Krippendorff2011ComputingKA}
Klaus Krippendorff. 2011.
\newblock \href {https://citeseerx.ist.psu.edu/document?repid=rep1&type=pdf&doi=de8e2c7b7992028cf035f8d907635de871ed627d} {Computing krippendorff's alpha-reliability}.

\bibitem[{Kuhn(1962)}]{Kuhn1962}
Thomas~S. Kuhn. 1962.
\newblock \emph{The Structure of Scientific Revolutions}.
\newblock University of Chicago Press, Chicago.

\bibitem[{Kummerfeld and Ramsey(2016)}]{10.1145/2939672.2939838}
Erich Kummerfeld and Joseph Ramsey. 2016.
\newblock \href {https://doi.org/10.1145/2939672.2939838} {Causal clustering for 1-factor measurement models}.
\newblock In \emph{Proceedings of the 22nd ACM SIGKDD International Conference on Knowledge Discovery and Data Mining}, KDD '16, page 1655–1664, New York, NY, USA. Association for Computing Machinery.

\bibitem[{Kıcıman et~al.(2023)Kıcıman, Ness, Sharma, and Tan}]{kıcıman2023causal}
Emre Kıcıman, Robert Ness, Amit Sharma, and Chenhao Tan. 2023.
\newblock \href {https://arxiv.org/abs/2305.00050} {Causal reasoning and large language models: Opening a new frontier for causality}.
\newblock \emph{Preprint}, arXiv:2305.00050.

\bibitem[{Lewis et~al.(2020)Lewis, Perez, Piktus, Petroni, Karpukhin, Goyal, K{\"u}ttler, Lewis, Yih, Rockt{\"a}schel et~al.}]{lewis2020retrieval}
Patrick Lewis, Ethan Perez, Aleksandra Piktus, Fabio Petroni, Vladimir Karpukhin, Naman Goyal, Heinrich K{\"u}ttler, Mike Lewis, Wen-tau Yih, Tim Rockt{\"a}schel, et~al. 2020.
\newblock Retrieval-augmented generation for knowledge-intensive nlp tasks.
\newblock \emph{Advances in Neural Information Processing Systems}, 33:9459--9474.

\bibitem[{Li et~al.(2024)Li, Panchendrarajan, and Zubiaga}]{li2024factfinderscheckthat2024refining}
Yufeng Li, Rrubaa Panchendrarajan, and Arkaitz Zubiaga. 2024.
\newblock \href {https://arxiv.org/abs/2406.18297} {Factfinders at checkthat! 2024: Refining check-worthy statement detection with llms through data pruning}.
\newblock \emph{Preprint}, arXiv:2406.18297.

\bibitem[{Lin et~al.(2024)Lin, Ravichander, Lu, Dziri, Sclar, Chandu, Bhagavatula, and Choi}]{lin2024the}
Bill~Yuchen Lin, Abhilasha Ravichander, Ximing Lu, Nouha Dziri, Melanie Sclar, Khyathi Chandu, Chandra Bhagavatula, and Yejin Choi. 2024.
\newblock \href {https://openreview.net/forum?id=wxJ0eXwwda} {The unlocking spell on base {LLM}s: Rethinking alignment via in-context learning}.
\newblock In \emph{The Twelfth International Conference on Learning Representations}.

\bibitem[{Liu et~al.(2024)Liu, Chen, Liu, Gong, Cheng, Han, and Zhang}]{liu2024discoveryhiddenworldlarge}
Chenxi Liu, Yongqiang Chen, Tongliang Liu, Mingming Gong, James Cheng, Bo~Han, and Kun Zhang. 2024.
\newblock \href {https://arxiv.org/abs/2402.03941} {Discovery of the hidden world with large language models}.
\newblock \emph{Preprint}, arXiv:2402.03941.

\bibitem[{Long et~al.(2022)Long, Schuster, and Pich{\'e}}]{long2022can}
Stephanie Long, Tibor Schuster, and Alexandre Pich{\'e}. 2022.
\newblock \href {https://openreview.net/forum?id=LQQoJGw8JD1} {Can large language models build causal graphs?}
\newblock In \emph{NeurIPS 2022 Workshop on Causality for Real-world Impact}.

\bibitem[{Meta(2024)}]{Llama31}
Meta. 2024.
\newblock \href {https://llama.meta.com/} {Meet llama 3.1}.

\bibitem[{Mohr et~al.(2022)Mohr, W{\"u}hrl, and Klinger}]{mohr-etal-2022-covert}
Isabelle Mohr, Amelie W{\"u}hrl, and Roman Klinger. 2022.
\newblock {C}o{VERT}: A corpus of fact-checked biomedical {COVID}-19 tweets.
\newblock In \emph{LREC:2022:1}, pages 244--257, Marseille, France. European Language Resources Association.

\bibitem[{Mooij et~al.(2016)Mooij, Peters, Janzing, Zscheischler, and Sch\"{o}lkopf}]{10.5555/2946645.2946677}
Joris~M. Mooij, Jonas Peters, Dominik Janzing, Jakob Zscheischler, and Bernhard Sch\"{o}lkopf. 2016.
\newblock Distinguishing cause from effect using observational data: methods and benchmarks.
\newblock \emph{J. Mach. Learn. Res.}, 17(1):1103–1204.

\bibitem[{Neal(2020)}]{neal2020causalitybook}
Brady Neal. 2020.
\newblock \href {https://www.bradyneal.com/Introduction_to_Causal_Inference-Dec17_2020-Neal.pdf} {\emph{Introduction to Causal Inference from a Machine Learning Perspective}}.

\bibitem[{OpenAI(2022)}]{chatgpt}
OpenAI. 2022.
\newblock \href {https://openai.com/index/chatgpt/} {Introducing chatgpt}.

\bibitem[{OpenAI(2024)}]{gpt4o}
OpenAI. 2024.
\newblock \href {https://openai.com/index/hello-gpt-4o/} {Hello gpt-4o}.

\bibitem[{Panchendrarajan and Zubiaga(2024)}]{Panchendrarajan_2024}
Rrubaa Panchendrarajan and Arkaitz Zubiaga. 2024.
\newblock \href {https://doi.org/10.1016/j.nlp.2024.100066} {Claim detection for automated fact-checking: A survey on monolingual, multilingual and cross-lingual research}.
\newblock \emph{Natural Language Processing Journal}, 7:100066.

\bibitem[{Pearl(2009)}]{pearl2009causality}
Judea Pearl. 2009.
\newblock \emph{Causality}.
\newblock Cambridge university press.

\bibitem[{Ross and Willson(2017)}]{Ross2017}
Amanda Ross and Victor~L. Willson. 2017.
\newblock \href {https://doi.org/10.1007/978-94-6351-086-8_4} {\emph{Paired Samples T-Test}}, pages 17--19.
\newblock SensePublishers, Rotterdam.

\bibitem[{Schmitz(2017)}]{schmitz2017predator}
Oswald Schmitz. 2017.
\newblock Predator and prey functional traits: understanding the adaptive machinery driving predator--prey interactions.
\newblock \emph{F1000Research}, 6.

\bibitem[{Shen et~al.(2020)Shen, Ma, Vemuri, and Simon}]{shen2020challenges}
Xinpeng Shen, Sisi Ma, Prashanthi Vemuri, and Gyorgy Simon. 2020.
\newblock Challenges and opportunities with causal discovery algorithms: application to alzheimer’s pathophysiology.
\newblock \emph{Scientific reports}, 10(1):2975.

\bibitem[{Shimizu et~al.(2006)Shimizu, Hoyer, Hyv{{\"a}}rinen, and Kerminen}]{shimizu06a}
Shohei Shimizu, Patrik~O. Hoyer, Aapo Hyv{{\"a}}rinen, and Antti Kerminen. 2006.
\newblock \href {http://jmlr.org/papers/v7/shimizu06a.html} {A linear non-gaussian acyclic model for causal discovery}.
\newblock \emph{Journal of Machine Learning Research}, 7(72):2003--2030.

\bibitem[{Si et~al.(2024)Si, Zhao, Zhu, Zhu, Lu, and Zhou}]{si-etal-2024-checkwhy}
Jiasheng Si, Yibo Zhao, Yingjie Zhu, Haiyang Zhu, Wenpeng Lu, and Deyu Zhou. 2024.
\newblock \href {https://doi.org/10.18653/v1/2024.acl-long.835} {{CHECKWHY}: Causal fact verification via argument structure}.
\newblock In \emph{ACL:2024:long}, pages 15636--15659, Bangkok, Thailand. acl.

\bibitem[{Silva et~al.(2006)Silva, Scheine, Glymour, and Spirtes}]{JMLR:v7:silva06a}
Ricardo Silva, Richard Scheine, Clark Glymour, and Peter Spirtes. 2006.
\newblock \href {http://jmlr.org/papers/v7/silva06a.html} {Learning the structure of linear latent variable models}.
\newblock \emph{Journal of Machine Learning Research}, 7(8):191--246.

\bibitem[{Spirtes et~al.(2000)Spirtes, Glymour, Scheines, and Heckerman}]{spirtes2000causation}
Peter Spirtes, Clark~N Glymour, Richard Scheines, and David Heckerman. 2000.
\newblock \emph{Causation, Prediction, and Search}.
\newblock MIT press.

\bibitem[{Wadden et~al.(2022)Wadden, Lo, Kuehl, Cohan, Beltagy, Wang, and Hajishirzi}]{wadden-etal-2022-scifact}
David Wadden, Kyle Lo, Bailey Kuehl, Arman Cohan, Iz~Beltagy, Lucy~Lu Wang, and Hannaneh Hajishirzi. 2022.
\newblock \href {https://doi.org/10.18653/v1/2022.findings-emnlp.347} {{S}ci{F}act-open: Towards open-domain scientific claim verification}.
\newblock In \emph{FINDINGS:2022:emnlp}, pages 4719--4734, Abu Dhabi, United Arab Emirates. acl.

\bibitem[{Wadhwa et~al.(2023)Wadhwa, Amir, and Wallace}]{wadhwa-etal-2023-revisiting}
Somin Wadhwa, Silvio Amir, and Byron Wallace. 2023.
\newblock \href {https://doi.org/10.18653/v1/2023.acl-long.868} {Revisiting relation extraction in the era of large language models}.
\newblock In \emph{ACL:2023:long}, pages 15566--15589, Toronto, Canada. acl.

\bibitem[{Wang and Shu(2023)}]{wang-shu-2023-explainable}
Haoran Wang and Kai Shu. 2023.
\newblock \href {https://doi.org/10.18653/v1/2023.findings-emnlp.416} {Explainable claim verification via knowledge-grounded reasoning with large language models}.
\newblock In \emph{FINDINGS:2023:emnlp}, pages 6288--6304, Singapore. acl.

\bibitem[{Wei et~al.(2022)Wei, Wang, Schuurmans, Bosma, brian ichter, Xia, Chi, Le, and Zhou}]{wei2022chain}
Jason Wei, Xuezhi Wang, Dale Schuurmans, Maarten Bosma, brian ichter, Fei Xia, Ed~H. Chi, Quoc~V Le, and Denny Zhou. 2022.
\newblock \href {https://openreview.net/forum?id=_VjQlMeSB_J} {Chain of thought prompting elicits reasoning in large language models}.
\newblock In \emph{Advances in Neural Information Processing Systems}.

\bibitem[{Yang et~al.(2022)Yang, Han, and Poon}]{yang2022survey}
Jie Yang, Soyeon~Caren Han, and Josiah Poon. 2022.
\newblock \href {https://link.springer.com/article/10.1007/s10115-022-01665-w} {A survey on extraction of causal relations from natural language text}.
\newblock \emph{Knowledge and Information Systems}, 64(5):1161--1186.

\bibitem[{Ze{\v{c}}evi{\'c} et~al.(2023)Ze{\v{c}}evi{\'c}, Willig, Dhami, and Kersting}]{ze2023causal}
Matej Ze{\v{c}}evi{\'c}, Moritz Willig, Devendra~Singh Dhami, and Kristian Kersting. 2023.
\newblock \href {https://openreview.net/forum?id=tv46tCzs83} {Causal parrots: Large language models may talk causality but are not causal}.
\newblock \emph{Transactions on Machine Learning Research}.

\bibitem[{Zhang et~al.(2011)Zhang, Peters, Janzing, and Sch\"{o}lkopf}]{10.5555/3020548.3020641}
Kun Zhang, Jonas Peters, Dominik Janzing, and Bernhard Sch\"{o}lkopf. 2011.
\newblock Kernel-based conditional independence test and application in causal discovery.
\newblock In \emph{Proceedings of the Twenty-Seventh Conference on Uncertainty in Artificial Intelligence}, UAI'11, page 804–813, Arlington, Virginia, USA. AUAI Press.

\bibitem[{Zhao et~al.(2024)Zhao, Deng, Yang, Wang, Zhang, Cheng, Lam, Shen, and Xu}]{10.1145/3674501}
Xiaoyan Zhao, Yang Deng, Min Yang, Lingzhi Wang, Rui Zhang, Hong Cheng, Wai Lam, Ying Shen, and Ruifeng Xu. 2024.
\newblock \href {https://doi.org/10.1145/3674501} {A comprehensive survey on relation extraction: Recent advances and new frontiers}.
\newblock \emph{ACM Comput. Surv.}, 56(11).

\bibitem[{Zhao et~al.(2023)Zhao, Yuan, Yuan, Deng, and Quan}]{zhao2023relation}
Youwen Zhao, Xiangbo Yuan, Ye~Yuan, Shaoxiong Deng, and Jun Quan. 2023.
\newblock \href {https://link.springer.com/article/10.1007/s13278-023-01095-8} {Relation extraction: advancements through deep learning and entity-related features}.
\newblock \emph{Social Network Analysis and Mining}, 13(1):92.

\bibitem[{Zheng et~al.(2018)Zheng, Aragam, Ravikumar, and Xing}]{10.5555/3327546.3327618}
Xun Zheng, Bryon Aragam, Pradeep Ravikumar, and Eric~P. Xing. 2018.
\newblock Dags with no tears: continuous optimization for structure learning.
\newblock In \emph{Proceedings of the 32nd International Conference on Neural Information Processing Systems}, NIPS'18, page 9492–9503, Red Hook, NY, USA. Curran Associates Inc.

\end{thebibliography}

\appendix
\section{Appendix}
\subsection{Related Work}
\label{sec:related_work}
\paragraph{Causal Discovery}
Causal discovery aims to uncover causal structures among variables, distinguishing itself from relation extraction in NLP by revealing novel causal relations rather than merely extracting known relations. While experimental approaches such as randomized controlled trials are gold standard methods\cite{fisher:1935}, practical limitations often necessitate statistical methods using observational data. These include constraint-based and score-based approaches \cite{spirtes2000causation, pearl2009causality, heckerman1995learning}. However, statistical methods face challenges in data collection and theoretical limitations. Recent advancements in LLMs have introduced new possibilities for causal discovery without direct data access \cite{kıcıman2023causal, ze2023causal, long2022can}. However, concerns about LLMs functioning as "causal parrots" and their ability to generalize to novel relations have been raised \cite{ze2023causal, feng2024pretrainingcorporalargelanguage}. Alternative approaches, such as using LLMs for variable proposer and combining them with statistical methods \cite{feng-etal-2023-less, feng-etal-2024-imo, liu2024discoveryhiddenworldlarge}, have emerged. Our work builds upon these ideas, introducing an automated document collection process, a hybrid causal discovery method integrating statistical and relation extraction techniques, and a hybrid approach for new variable proposal.

\paragraph{Relation Extraction}
Relation extraction aims to transform unstructured textual relations into structured relation tuples of the form $<e_1, r, e_2>$, where $e_1$ and $e_2$ represent entities and $r$ denotes the relation between them \cite{yang2022survey, dasgupta-etal-2018-automatic-extraction}. While relation extraction can identify cause-effect relationships from documents, it fundamentally differs from causal discovery in that it relies on explicitly stated relations in texts, whereas causal discovery can uncover novel causal relationships from observational data even in the absence of explicit textual mentions. Nevertheless, relation extraction can serve as a complementary method for identifying commonly known causal relations in textual data. Several studies have focused on extracting causal relations from natural language texts \cite{balashankar-etal-2019-identifying, bui2010extracting, CHANG2006662, feng-etal-2025-causalscore}. The methods for causality extraction can be divided into pattern-based and deep learning-based approaches. Pattern-based methods utilize predefined linguistic patterns to extract relevant text segments, which are then converted into tuples using hand-crafted algorithms \cite{10.5555/645359.651023, khoo-etal-2000-extracting}. However, these methods often suffer from limited coverage of causal relations and require significant effort in pattern design. Deep learning-based methods employ neural networks to learn high-level abstract features and representations from sentences, framing relation extraction as a sequence-to-sequence task \cite{zhao2023relation, 10.1145/3674501}. While these approaches offer improved performance, they typically require large fine-tuning datasets and may not consistently produce structurally correct output tuples.

A notable limitation of many relation extraction systems is the lack of verification for extracted relations, potentially leading to the extraction of false or unreliable relations from untrustworthy sources \cite{si-etal-2024-checkwhy,wadhwa-etal-2023-revisiting}. Our work addresses this issue by adopting a novel approach: instead of directly extracting causal relations from documents, we pre-create textual mentions of causal relations (e.g., "smoking causes lung cancer") and employ LLMs to verify the veracity of these relations based on relevant documents. We consider a causal relation to hold if the majority of documents support its veracity, thereby enhancing the reliability of our extracted causal relations.

\paragraph{Claim Verification}
Claim verification aims to assess the veracity of claims based on relevant documents \cite{bekoulis2021review}. This process typically encompasses several key components: claim detection, document retrieval, veracity prediction, and explanation generation. Research in this field often focuses on specific aspects of the verification pipeline. For instance, \citet{Panchendrarajan_2024} and \citet{li2024factfinderscheckthat2024refining} concentrate on identifying check-worthy statements from large text corpora. Others, such as \citet{wadden-etal-2022-scifact} and \citet{mohr-etal-2022-covert}, prioritize veracity prediction, while \citet{wang-shu-2023-explainable} emphasize the importance of generating explanations for verification outcomes. The emergence of LLMs has significantly influenced the field, with numerous studies leveraging LLMs for claim verification through carefully crafted prompts \cite{kim2024llmsproducefaithfulexplanations, bazaga2024unsupervised, asai2024selfrag}. Building on these advancements, one branch of our hybrid causal discovery approach reframes causal discovery as a causal relation verification task. We employ LLMs to assess the veracity of causal relations based on retrieved documents, subsequently incorporating verified relations into a causal graph. This methodology bridges the gap between traditional claim verification techniques and causal discovery, offering a novel approach to uncovering and validating causal relations.

\subsection{Reproducibility Statement}
We release our code and scripts at \url{https://github.com/WilliamsToTo/iris}. Detailed descriptions of the algorithms used in each component of our framework can be found in Appendix~\ref{apx:algorithms}. We provide all prompts utilized throughout our framework in Appendix~\ref{apx:prompts}. The ground-truth causal graphs employed in our evaluation experiments are outlined in Appendix~\ref{apx:ground_truth_causal_graphs}. Additionally, Appendix~\ref{apx:causal_relation_annotation_task} presents human annotation instruction and interface for the human annotation tasks involved in evaluating the expanded causal graphs. The annotated expanded causal graphs, alongside the predicted causal graphs, are documented in Appendix~\ref{apx:expanded_causal_graphs}.

\subsection{Algorithms}
\label{apx:algorithms}
In this section, we provide detailed descriptions of the algorithms for each component of our method. The data collection and value extraction process is outlined in \Algref{alg:data_collect_extract}. The hybrid causal discovery algorithm can be found in \Algref{alg:hybrid_causal_discovery}. Finally, the algorithm for proposing missing variables is detailed in \Algref{alg:missing_variable_proposal}. 

\begin{algorithm}[ht]
\caption{Document Collection and Value Extraction}
\label{alg:data_collect_extract}
\begin{algorithmic}
\Require Initial Variables $\sZ$, LLM $\mM$, threshold $T$, prompt $l$

\State \textbf{Document Collection}
\State $\sD \gets \emptyset$ \Comment{Initialize an empty set for collected documents}
\While{$|\sD| < T$}
    \State $queries \gets [(\rz_1, \rz_2, \ldots, \rz_n), (\rz_1, \rz_2, \ldots, \rz_{n-1}), \ldots, (\rz_i)]$ \\ \Comment{queries considering all variables and their synonyms}
    \For{each $q$ in $queries$}
        \State $n \gets 20 \times \text{len}(q)$ \Comment{Determine the number of URLs to collect}
        \State $urls \gets \text{google\_search}(q, n)$ \Comment{Search with query $q$ and retrieve top-$n$ URLs}
        \For{each $url$ in $urls$}
            \State $D \gets$ \text{extract text from } $url$ 
            \State $\sD \gets \sD \cup \{D\}$ \Comment{Add extracted text to the document set}
        \EndFor
    \EndFor
\EndWhile

\State
\State \textbf{Value Extraction}
\State $\mV \gets \text{Matrix of dimensions } T \times N$ \Comment{Initialize a matrix with $T$ rows and $N$ columns}
\For{each $d_i$ in $\sD$}
    \For{each $\rz_j$ in $\sZ$}
        \State $o_{ij} \gets \mM(l(d_i, \rz_j))$ \Comment{Determine value of $\rz_j$ in $d_i$ by LLM}
        \State $v_{ij} \gets \text{extract}(o_{ij})$ \Comment{Extract value from LLM output}
        \State $\mV[i][j] \gets v_{ij}$ \Comment{Store the value $v_{ij}$ in matrix $\mV$ at position $(i, j)$}
    \EndFor
\EndFor
\State \textbf{Output: } $\sD, \mV$
\end{algorithmic}
\end{algorithm}

\begin{algorithm}[ht]
\caption{Hybrid Causal Discovery}
\label{alg:hybrid_causal_discovery}
\begin{algorithmic}
\Require Initial variables $\sZ$, LLM $\mM$, structured data $\sX$, prompt $l$, hyperparameters $\alpha, \beta$

\State \textbf{Statistical Causal Discovery}
\State $\hat{\gG_{s}} \gets \text{causal\_discovery\_alg}(\sX)$ \Comment{Apply  causal discovery algorithms (e.g., PC algorithm)}

\State
\State \textbf{Causal Relation Verification}
\State $\hat{\gG_{v}} \gets$ \text{causal graph with no edges}
\State $\text{remove\_edges} \gets \emptyset$
\For{each $\rz_i$ in $\sZ$}
    \For{each $\rz_j$ in $\sZ$}
        \If {$\rz_i \neq \rz_j$}
            \State $r \gets $ "$\rz_i$ causes $\rz_j$" 
            \State $veracity_r \gets \emptyset$ \Comment{Initialize the veracity list for relation $r$}
            \For{each $d$ in $\sD_{\rz_i, \rz_j}$} \Comment{$\sD_{\rz_i, \rz_j}$ denotes documents containing both $\rz_i$ and $\rz_j$}
                \State $ver_{d} \gets \mM(l(r, d))$ \Comment{Determine the veracity of $r$ based on document $d$}
                \State $veracity_r \gets veracity_r \cup \{ver_{d}\}$ 
            \EndFor
            \If {$veracity_r.count(True) > \alpha \times len(veracity_r)$}
                \State $\hat{\gG_{v}} \gets \hat{\gG_{v}} \cup \{r\}$ \Comment{Add relation $r$ to the causal graph $\hat{\gG_{v}}$}
            \ElsIf {$veracity_r.count(False) > \beta \times len(veracity_r)$}
                \State $\text{remove\_edges} \gets \text{remove\_edges} \cup \{r\}$ 
            \EndIf
        \EndIf
    \EndFor
\EndFor

\State
\State \textbf{Merge $\hat{\gG_{s}}$ and $\hat{\gG_{v}}$}
\For{each edge $r$ in $\hat{\gG_{v}}$}
    \State $\hat{\gG_{s}} \gets \hat{\gG_{s}} \cup \{r\}$ \Comment{Add relation $r$ to $\hat{\gG_{s}}$}
\EndFor
\For{each edge $r$ in $\text{remove\_edges}$}
    \State $\hat{\gG_{s}} \gets \hat{\gG_{s}} \setminus \{r\}$ \Comment{Remove relation $r$ from $\hat{\gG_{s}}$ if it exists}
\EndFor
\State $\hat{\gG} \gets \hat{\gG_{s}}$ \Comment{The final merged causal graph}
\State \textbf{Output:} $\hat{\gG}$
\end{algorithmic}
\end{algorithm}
\begin{algorithm}[ht]
\caption{Missing Variable Proposal}
\label{alg:missing_variable_proposal}
\begin{algorithmic}
\Require Initial variables $\sZ$, LLM $\mM$, collected documents $\sD$, prompt $l$, hyperparameter $\alpha$

\State \textbf{Step 1: Abstract Missing Variable Candidates}
\State $\sZ_{c} \gets \emptyset$ \Comment{Initialize the set of candidates}
\For{each document $d$ in $\sD$}
    \State $\rz \gets \mM(l(\sZ, d))$ \Comment{Abstract a candidate variable from document $d$}
    \State $\sZ_{c} \gets \sZ_{c} \cup \{\rz\}$ 
\EndFor

\State
\State \textbf{Step 2: Missing Variable Proposal Based on Verified Causal Relations}
\State $\sZ_{m} \gets \emptyset$ \Comment{Initialize the set of missing variables}
\For{each variable $\rz_i$ in $\sZ_{c}$}
    \For{each given variable $\rz_j$ in $\sZ$}
        \State $r_1 \gets \text{"$\rz_i$ causes $\rz_j$"}$
        \State $veracity_{r_1} \gets \emptyset$ \Comment{Initialize the veracity list for relation $r_1$}
        \For{each document $d$ in $\sD_{\rz_i, \rz_j}$} \Comment{$\sD_{\rz_i, \rz_j}$ denotes documents containing both $\rz_i$ and $\rz_j$}
            \State $ver_{d} \gets \mM(l(r_1, d))$ \Comment{Determine the veracity of $r1$ based on document $d$}
            \State $veracity_{r_1} \gets veracity_{r_1} \cup \{ver_{d}\}$ 
        \EndFor
        \If {$\text{veracity}_{r_1}.count(\text{True}) > \alpha \times \text{veracity}_{r_1}.count(\text{False})$}
            \State $\sZ_{m} \gets \sZ_{m} \cup \{\rz_i\}$ \Comment{Add $\rz_i$ to the set of proposed variables}
        \EndIf

        \State $r_2 \gets \text{"$\rz_j$ causes $\rz_i$"}$ \Comment{Repeat the process for the reverse causal relation}
        % \State \ldots
    \EndFor
\EndFor

\State
\State \textbf{Step 3: Missing Variable Proposal Based on Statistical Methods}
\State $\sS \gets \emptyset$ \Comment{Initialize a set for PMI scores}
\For{each variable $\rz_i$ in $\sZ_{c}$}
    \State $s_i \gets \emptyset$ 
    \For{each given variable $\rz_j$ in $\sZ$}
        \State $s_{ij} \gets \text{PMI}(\rz_i, \rz_j)$ \Comment{Compute PMI of $(\rz_i, \rz_j)$ by \Eqref{equ:pmi}}
        \State $s_i \gets s_i \cup \{s_{ij}\}$ 
    \EndFor
    \State $\sS \gets \sS \cup \{\sum(s_i)\}$ \Comment{Aggregate the PMI scores for $\rz_i$}
\EndFor
\State $\sZ_{m} \gets \sZ_{m} \cup \text{top-k}(\sS, \sZ_{c})$ \Comment{PMI scores}

\State \textbf{Output:} $\sZ_{m}$ 
\end{algorithmic}
\end{algorithm}

\subsection{Prompt Engineering Details}
\label{apx:prompts}
Prompts were designed using different strategies and ultimately adopted the chain-of-thought (CoT) \cite{wei2022chain} prompting approach, as shown in Table~\ref{tab:prompt_iris}. These prompts contain retrieved document, task descriptions and stepwise instructions to complete tasks. Then we require LLMs to output final answer with specific format to easily extract answers. We also tried zero-shot prompts, but it demonstrated poor performance for value extraction, causal relation extraction, and missing variable abstraction. Few-shot prompts often exceeded the maximum input length for LLMs, as they required incorporating multiple long documents into the prompt. In contrast, CoT prompting provided better performance in all components. 

To demonstrate that IRIS is LLM-agnostic, we use the same prompt across all LLMs during the evaluation. In our study, we use a separate validation set (not overlapping with test data) to compare prompts. This validation set is built using high-confident causal relations from CauseNet \cite{10.1145/3340531.3412763} with randomly paired non-causal relations. For prompt selection, we first manually write a pool of manually written prompts from different researchers, then use GPT-4 to refine these prompts. We evaluate all human and LLM-refined prompts on the validation set and select the prompt that has the best performance (highest F1).

For the evaluation of the whole framework, the prompt used in the 0-shot, CoT, and RAG baselines is shown in Table~\ref{tab:prompt_expanded_causal_graphs_baseline}. For the evaluation of the missing variable proposal, the prompt used the 0-shot, CoT, and RAG baselines is shown in Table~\ref{tab:prompt_missing_variable_proposal_baseline}.

\begin{table*}[t]
\centering
\resizebox{\textwidth}{!}{%
\begin{tabular}{l}
\hline
\multicolumn{1}{c}{\textbf{Value Extraction}}                                                                                                                                                                                                                                                                                                                                                                                                                                                                                                                                                                                                                                                                                                                                                                                                                                                                                                           \\
\begin{tabular}[c]{@{}l@{}}Given a document: \{doc\}\\ \\ Please complete the below task.\\ We have a variable named '\{var\}'. The value of variable '\{var\}' is True or False.\\ True indicates that the existence of '\{var\}' can be inferred from the document, whereas False suggests that the existence of '\{var\}' cannot be inferred from this document.\\ Based on the document provided, what is the most appropriate value for '\{var\}' that can be inferred?\\ Please form the answer using the following format.\\ First, provide an introductory sentence that explains what information will be discussed.\\ Next, list generated answer in detail, ensuring clarity and precision.\\ Finally, conclude the final answer of the inferred value for '\{var\}' using the following template:\\ The value of '\{var\}' is \_\_\_\_.\end{tabular}                                                                                        \\ \hline
\multicolumn{1}{c}{\textbf{Causal Relation Verification}}                                                                                                                                                                                                                                                                                                                                                                                                                                                                                                                                                                                                                                                                                                                                                                                                                                                                                               \\
\begin{tabular}[c]{@{}l@{}}Given a document: \{doc\}\\ \\ Please complete the below task.\\ We have a claim: '\{claim\}'.  We need to check the veracity of this claim. The value of veracity is True or False or Unknown.\\ True indicates that the given document supports this claim, \\ False indicates that the given document refutes the claim.\\ Unknown indicates that the given document has no relation to the claim.\\ Please form the answer with a logical reasoning chain according to the following format.\\ First, provide an introductory sentence that explains what information will be discussed. \\ Next, list the logical reasoning chain in detail, ensuring clarity and precision.\\ Finally, conclude the veracity of claim '\{claim\}' using the following template:\\ The veracity of claim '\{claim\}' is \_\_\_.\end{tabular}                                                                                            \\ \hline
\multicolumn{1}{c}{\textbf{Missing Variable Abstraction}}                                                                                                                                                                                                                                                                                                                                                                                                                                                                                                                                                                                                                                                                                                                                                                                                                                                                                               \\
\begin{tabular}[c]{@{}l@{}}Given a document: \{doc\}\\ \\ Please complete the below task.\\ We have some given variables: '\{initial\_variables\}'.\\ What are the high-level variables in the provided document that have causal relations to variables in the given variable set?\\ Please form the answer using the following format.\\ First, propose as many variables as possible that have causal relationships with the given variables, based on your understanding of the document. \\ Please ensure these proposed variables are different from the ones already provided.\\ Next, refine your list of candidate variables by reducing semantic overlap among them and shortening their names for clarity.\\ Finally, determine the most reliable variable candidate as the final answer using the template provided below:\\ The new abstracted variable is \textless{}var\textgreater{}\_\_\_\_\textless{}/var\textgreater{}.\end{tabular} \\ \hline
\end{tabular}%
}
\caption{The prompts used in IRIS, where doc indicates the content of a document, claim refers to a causal relation (\eg smoking causes lung cancer).}
\label{tab:prompt_iris}
\end{table*}

\begin{table*}[t]
\centering
% \resizebox{\textwidth}{!}{%
\begin{tabular}{l}
\hline
\multicolumn{1}{c}{\textbf{0-shot}}                                                                                                                                                                                                                                                                                                                                                                                                                                                                                                                                                                                                                                                                                                                                                                                                                                                                                                                                       \\
\begin{tabular}[c]{@{}l@{}}The task is to determine the cause-effect relation between two variables.\\ The variables are: {variable1} and {variable2}.\\ Your answer should be one of the following:\\ {variable1} → {variable2} (if {variable1} causes {variable2})\\ {variable1} ← {variable2} (if {variable2} causes {variable1})\\ No causal relation (if there is no clear cause-effect relationship)\\ \\ Let's provide a step-by-step process to analyze the relation between them,  \\ then provide your final answer using the following format:\\ The final answer is: {variable1} → {variable2} or {variable1} ← {variable2} or No causal relation\end{tabular}                                                                                                                                                                                                                                                                                               \\ \hline
\multicolumn{1}{c}{\textbf{CoT}}                                                                                                                                                                                                                                                                                                                                                                                                                                                                                                                                                                                                                                                                                                                                                                                                                                                                                                                                          \\
\begin{tabular}[c]{@{}l@{}}The task is to determine the cause-effect relation between two variables.\\ The variables are:  {variable1} and {variable2}.\\ Your answer should be one of the following:\\ {variable1} → {variable2} (if {variable1} causes {variable2})\\ {variable1} ← {variable2} (if {variable2} causes {variable1})\\ No causal relation (if there is no clear cause-effect relationship)\\ \\ Let's analyze the relation through the following steps:\\ First, briefly describe each variable and its typical behavior.\\ Second, does one variable naturally precede the other in time or logic?\\ Third, are there common confounders or external factors that could explain the relationship?\\ Finally, provide your final answer in the following format:\\ The final answer is: {variable1} → {variable2} or {variable1} ← {variable2} or No causal relation\end{tabular}                                                                        \\ \hline
\multicolumn{1}{c}{\textbf{RAG}}                                                                                                                                                                                                                                                                                                                                                                                                                                                                                                                                                                                                                                                                                                                                                                                                                                                                                                                                          \\
\begin{tabular}[c]{@{}l@{}}Analyze the relevant information from the retrieved document: \\ {doc}\\ The task is to determine the cause-effect relation between two variables.\\ The variables are:  {variable1} and {variable2}.\\ Your answer should be one of the following:\\ {variable1} → {variable2} (if {variable1} causes {variable2})\\ {variable1} ← {variable2} (if {variable2} causes {variable1})\\ No causal relation (if there is no clear cause-effect relationship)\\ \\ Let's analyze the relation through the following steps:\\ First, briefly describe each variable and its typical behavior.\\ Second, does one variable naturally precede the other in time or logic?\\ Third, does the retrieved document provide any information that explains the relationship?\\ Finally, provide your final answer in the following format:\\ The final answer is: {variable1} → {variable2} or {variable1} ← {variable2} or No causal relation\end{tabular} \\ \hline
\end{tabular}%
% }
\caption{The prompt used in the baselines for evaluation of expanded causal graphs.}
\label{tab:prompt_expanded_causal_graphs_baseline}
\end{table*}

\begin{table*}[t]
\centering
\resizebox{\textwidth}{!}{%
\begin{tabular}{l}
\hline
\multicolumn{1}{c}{\textbf{0-shot}}                                                                                                                                                                                                                                                                                                                                                                                                                                                                                                                                                                                                                                                                                                                                     \\
\begin{tabular}[c]{@{}l@{}}The task is to identify new variables that are causally related to the given variables.  \\ Given variables: {variables}  \\ Follow a step-by-step approach to analyze the given variables and determine relevant causal relationships. \\ Then, present your final answer in the following format:  \\ Proposed variables: [variable1, variable2, ...]\end{tabular}                                                                                                                                                                                                                                                                                                                                                                         \\ \hline
\multicolumn{1}{c}{\textbf{CoT}}                                                                                                                                                                                                                                                                                                                                                                                                                                                                                                                                                                                                                                                                                                                                        \\
\begin{tabular}[c]{@{}l@{}}The task is to identify new variables that are causally related to the given variables.\\ Given variables: {variables}\\ Let's break this down step by step to systematically analyze the given variables and determine relevant causal relationships:\\ First, understand the given variables.\\ Second, propose potential direct causes and effects associated with the given variables.\\ Third,  verify that the proposed variables align with real-world causal structures.\\ Finally, present your final answer in the following format:  \\ Proposed variables: [variable1, variable2, ...]\end{tabular}                                                                                                                              \\ \hline
\multicolumn{1}{c}{\textbf{RAG}}                                                                                                                                                                                                                                                                                                                                                                                                                                                                                                                                                                                                                                                                                                                                        \\
\begin{tabular}[c]{@{}l@{}}Given a document: {doc}\\ The task is to identify new variables that are causally related to the given variables.\\ Given variables: {variables}\\ Let's break this down step by step to systematically analyze the given variables and determine relevant causal relationships:\\ First, understand the given variables.\\ Second, propose potential direct causes and effects associated with the given variables based on the information in the given document.\\ Third,  verify that the proposed variables align with real-world causal structures.\\ Finally, present the most reliable variable candidates as the final proposed variables in the following format:  \\ Proposed variables: [variable1, variable2, ...]\end{tabular} \\ \hline
\end{tabular}%
}
\caption{The prompt used in the baselines to evaluate the missing variable proposal.}
\label{tab:prompt_missing_variable_proposal_baseline}
\end{table*}

\subsection{Causal Relation Annotation Task for Expanded Variables}
\label{apx:causal_relation_annotation_task}
The detailed instructions for the causal relation annotation task for expanded variables are presented in Table~\ref{tab:causal_relation_annotation_instruction}. This table provides comprehensive guidance to annotators on how to identify and annotate causal relations among the given variables.

% Please add the following required packages to your document preamble:
% \usepackage{graphicx}
% \usepackage[normalem]{ulem}
% \useunder{\uline}{\ul}{}
\begin{table*}[t]
\centering
\resizebox{\textwidth}{!}{%
\begin{tabular}{l}
\hline
\textbf{Causal Relation Annotation Task}                                                                                                                  \\
                                                                                                                                                          \\
\textbf{Task overview:}                                                                                                                                   \\
Your task is to identify and annotate causal relations among a set of variables. A causal relation exists when one variable directly influences another.  \\
                                                                                                                                                          \\
\textbf{Instructions:}                                                                                                                                    \\
1. Consider each pair of variables and determine if there is a direct causal relationship between them.                                                   \\
2. If you believe variable A causes variable B, indicate this as: A → B                                                                                   \\
3. Be cautious of confusing correlation with causation. Only mark a relationship if you believe there is a direct causal link.                            \\
4. Consider the direction of causality carefully. For example, "Obesity → Heart Failure" suggests obesity causes heart failure, not the other way around. \\
5. It's okay to have multiple causes for a single effect, or multiple effects from a single cause.                                                        \\
6. Not all variables will necessarily have causal relationships with others.                                                                              \\
7. Use your best judgment based on available knowledge and logical reasoning.                                                                             \\
                                                                                                                                                          \\
\textbf{Examples:}                                                                                                                                        \\
lifestyle -> obesity                                                                                                                                      \\
heart defect -> cardiac output                                                                                                                            \\
genetic disorder -> heart defect                                                                                                                          \\
                                                                                                                                                          \\
\textbf{Submission:}                                                                                                                                      \\
Please submit your annotations as a list of causal relations in the format: Variable A -> Variable B                                                      \\
Thank you for your careful consideration of this task!                                                                                                    \\
                                                                                                                                                          \\
                                                                                                                                                          \\
                                                                                                                                                          \\
\textbf{Task 1: Cancer}                                                                                                                                   \\
                                                                                                                                                          \\
\textbf{Variables:}                                                                                                                                       \\
pollution                                                                                                                                                 \\
smoker                                                                                                                                                    \\
cancer                                                                                                                                                    \\
x-ray                                                                                                                                                     \\
dyspnoea                                                                                                                                                  \\
air quality                                                                                                                                               \\
education                                                                                                                                                 \\
health issues                                                                                                                                             \\
toxicity                                                                                                                                                  \\
chronic illness                                                                                                                                           \\
covid-19                                                                                                                                                  \\
inflammation                                                                                                                                              \\
respiratory issues                                                                                                                                        \\
immunity                                                                                                                                                  \\
carcinogens                                                                                                                                               \\
early detection                                                                                                                                           \\
                                                                                                                                                          \\
\textbf{Causal Relations:}                                                                                                                                \\
...                                                                                                                                                       \\
                                                                                                                                                          \\ \hline
\end{tabular}%
}
\caption{Instructions and interface of causal relation annotation task.}
\label{tab:causal_relation_annotation_instruction}
\end{table*}

\subsection{Human-annotated Causal Graphs for Expanded Variables}
\label{apx:expanded_causal_graphs}
\begin{figure*}[h]
\centering
\begin{subfigure}{0.8\textwidth}
    \includegraphics[width=\linewidth]{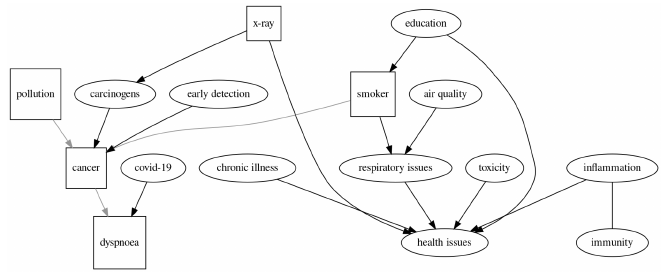}
    \caption{\ourmethod}
    \label{fig:predicted_expanded_cancer}
\end{subfigure}
\begin{subfigure}{0.8\textwidth}
    \includegraphics[width=\linewidth]{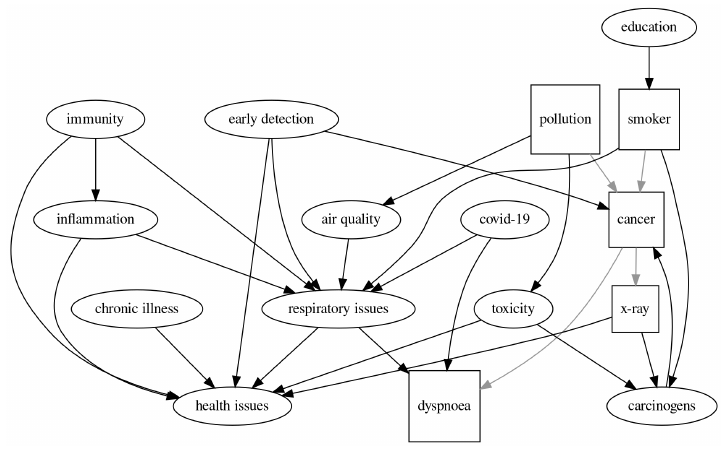}
    \caption{Human}
    \label{fig:human_annotation_expanded_cancer}
\end{subfigure}
\caption{Illustration of expanded causal graphs for Cancer. Squared nodes represent initial variables, while round nodes denote new proposed variables.}
\label{fig:expanded_causal_graphs_cancer}
\end{figure*}

\begin{figure*}[h]
\centering
\begin{subfigure}{0.8\textwidth}
    \includegraphics[width=\linewidth]{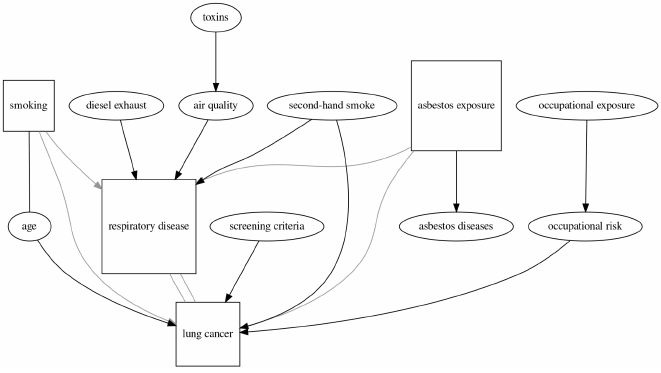}
    \caption{\ourmethod}
    \label{fig:predicted_expanded_respiratory_disease}
\end{subfigure}
\begin{subfigure}{0.8\textwidth}
    \includegraphics[width=\linewidth]{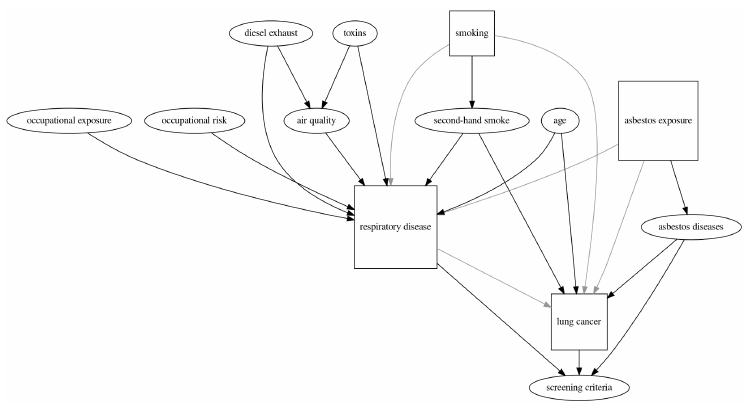}
    \caption{Human}
    \label{fig:human_annotation_expanded_respiratory_disease}
\end{subfigure}
\caption{Illustration of expanded causal graphs for Respiratory Disease. Squared nodes represent initial variables, while round nodes denote new proposed variables.}
\label{fig:expanded_causal_graphs_respiratory_disease}
\end{figure*}

\begin{figure*}[h]
\centering
\begin{subfigure}{0.8\textwidth}
    \includegraphics[width=\linewidth]{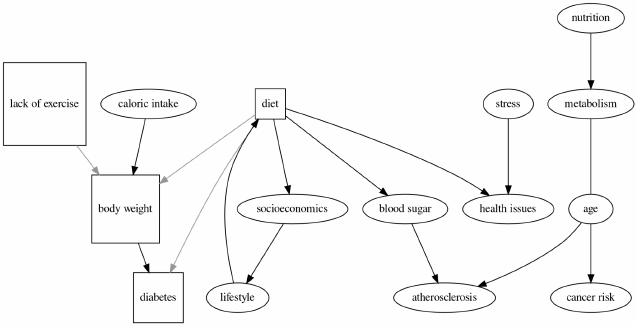}
    \caption{\ourmethod}
    \label{fig:predicted_expanded_diabetes}
\end{subfigure}
\begin{subfigure}{0.8\textwidth}
    \includegraphics[width=\linewidth]{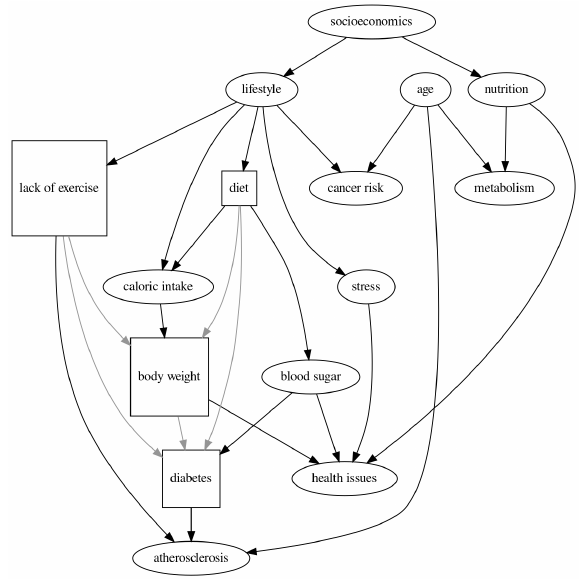}
    \caption{Human}
    \label{fig:human_annotation_expanded_diabetes}
\end{subfigure}
\caption{Illustration of expanded causal graphs for Diabetes. Squared nodes represent initial variables, while round nodes denote new proposed variables.}
\label{fig:expanded_causal_graphs_diabetes}
\end{figure*}

\begin{figure*}[h]
\centering
\begin{subfigure}{0.7\textwidth}
    \includegraphics[width=\linewidth]{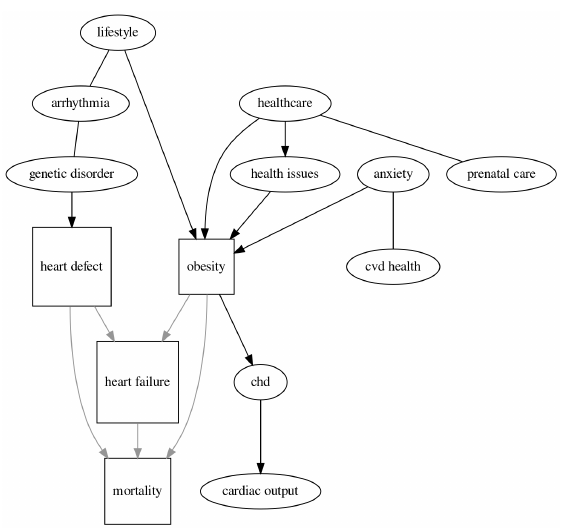}
    \caption{\ourmethod}
    \label{fig:predicted_expanded_obesity}
\end{subfigure}
\begin{subfigure}{0.7\textwidth}
    \includegraphics[width=\linewidth]{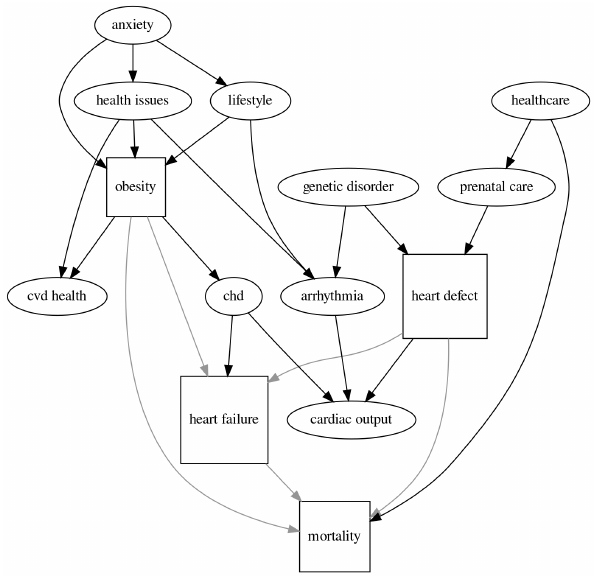}
    \caption{Human}
    \label{fig:human_annotation_expanded_obesity}
\end{subfigure}
\caption{Illustration of expanded causal graphs for Obesity. Squared nodes represent initial variables, while round nodes denote new proposed variables.}
\label{fig:expanded_causal_graphs_obesity}
\end{figure*}

\begin{figure*}[h]
\centering
\begin{subfigure}{0.6\textwidth}
    \includegraphics[width=\linewidth]{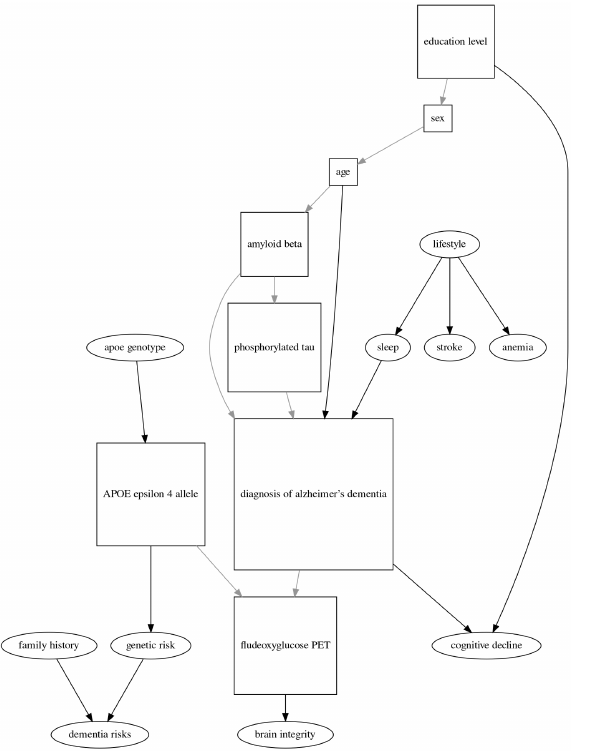}
    \caption{\ourmethod}
    \label{fig:predicted_expanded_adni}
\end{subfigure}
\begin{subfigure}{0.6\textwidth}
    \includegraphics[width=\linewidth]{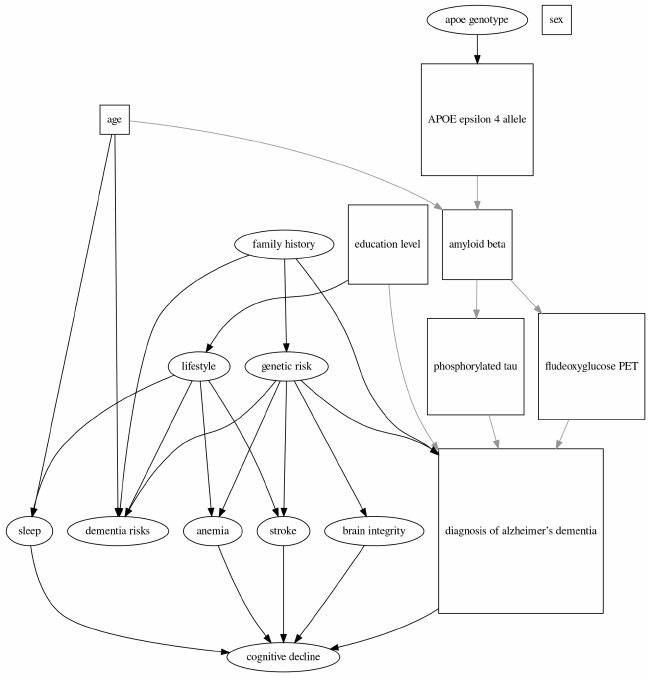}
    \caption{Human}
    \label{fig:human_annotation_expanded_adni}
\end{subfigure}
\caption{Illustration of expanded causal graphs for ADNI. Squared nodes represent initial variables, while round nodes denote new proposed variables.}
\label{fig:expanded_causal_graphs_adni}
\end{figure*}

\begin{figure*}[h]
\centering
\begin{subfigure}{0.65\textwidth}
    \includegraphics[width=\linewidth]{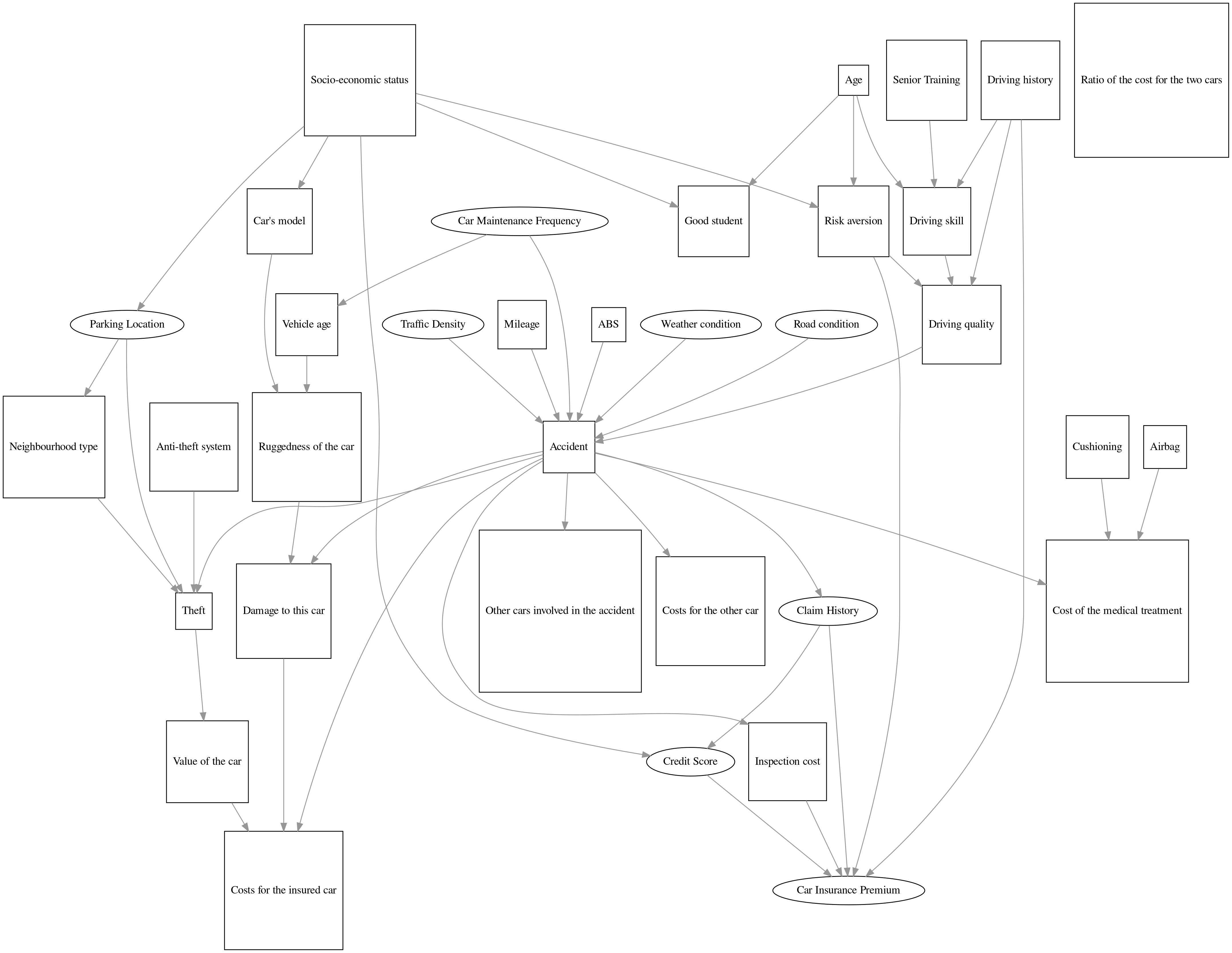}
    \caption{\ourmethod}
    \label{fig:predicted_expanded_insurance}
\end{subfigure}
\begin{subfigure}{0.6\textwidth}
    \includegraphics[width=\linewidth]{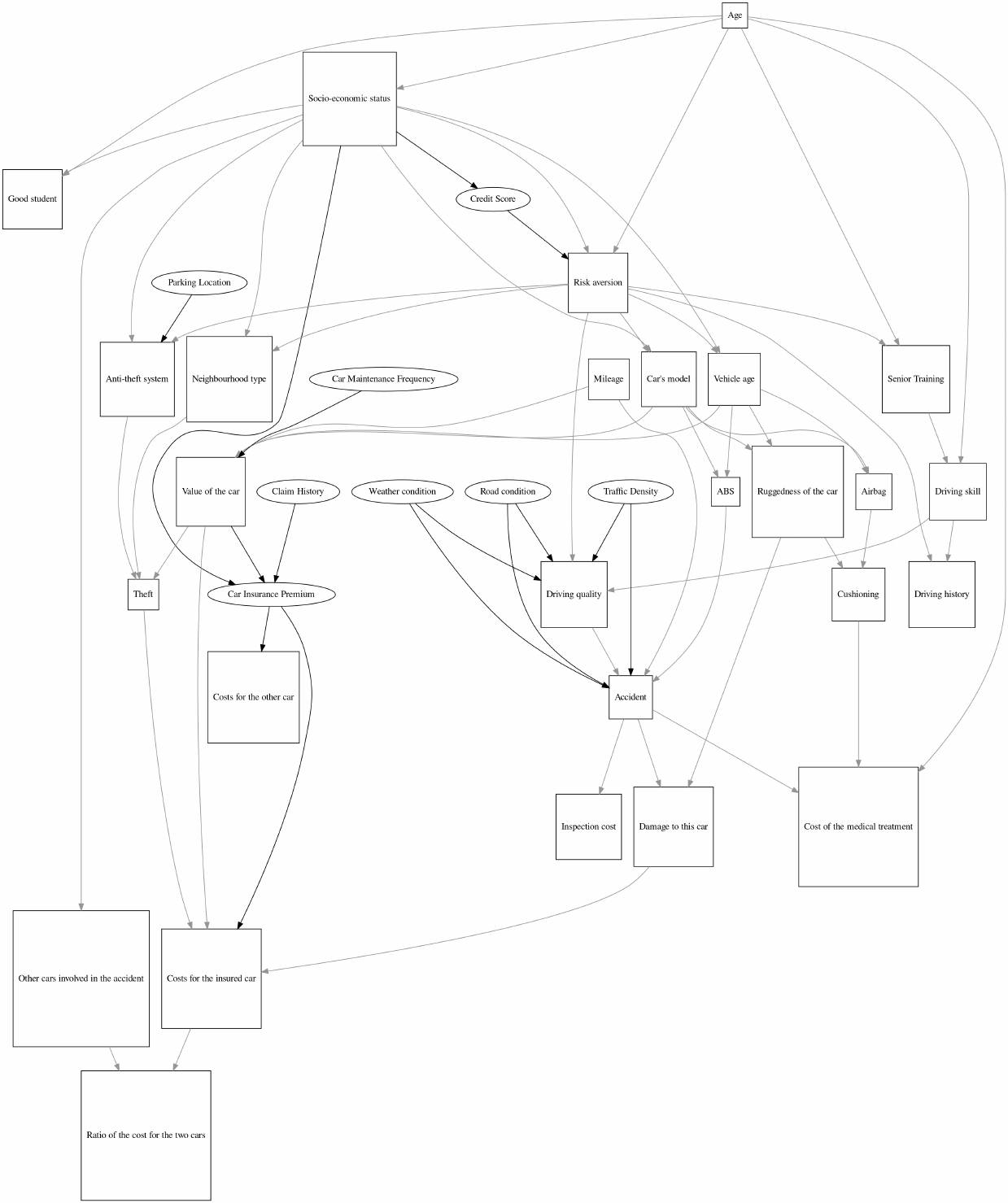}
    \caption{Human}
    \label{fig:human_annotation_expanded_insurance}
\end{subfigure}
\caption{Illustration of expanded causal graphs for Insurance.}
\label{fig:expanded_causal_graphs_insurance}
\end{figure*}

The human-annotated causal graphs for expanded variables are demonstrated in Figure \ref{fig:expanded_causal_graphs_cancer}, \ref{fig:expanded_causal_graphs_respiratory_disease}, \ref{fig:expanded_causal_graphs_diabetes}, \ref{fig:expanded_causal_graphs_obesity}, \ref{fig:expanded_causal_graphs_adni}, \ref{fig:expanded_causal_graphs_insurance}. 
The statistics of human-annotated causal graphs is presented in Table~\ref{tab:statistics_expanded_graphs}.

\begin{table}[t]
\centering
% \resizebox{\linewidth}{!}{%
\begin{tabular}{l|cc}
\hline
Causal Graph        & Node & Edge \\ \hline
Cancer              & 16   & 28   \\
Respiratory Disease & 13   & 22   \\
Diabetes            & 15   & 26   \\
Obesity             & 14   & 25   \\
ADNI                & 18   & 27   \\
Insurance           & 35   & 67   \\ \hline
\end{tabular}%
% }
\caption{Statistics of human-annotated causal graph for expanded variables.}
\label{tab:statistics_expanded_graphs}
\end{table}

\subsection{Ground-Truth Causal Graphs of Initial Variables}
\label{apx:ground_truth_causal_graphs}
\begin{figure*}[ht]
\centering

\begin{subfigure}{0.32\textwidth}
    \includegraphics[width=\linewidth]{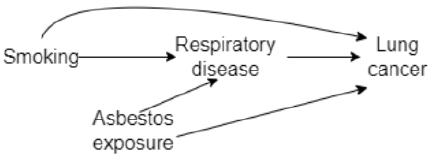}
    \caption{Respiratory Disease}
    \label{fig:true_respiratory_disease}
\end{subfigure}
\hfill
\begin{subfigure}{0.32\textwidth}
    \includegraphics[width=\linewidth]{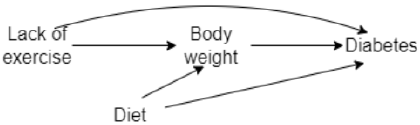}
    \caption{Diabetes}
    \label{fig:true_diabetes}
\end{subfigure}
\hfill
\begin{subfigure}{0.32\textwidth}
    \includegraphics[width=\linewidth]{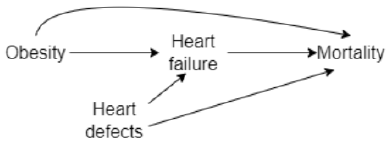}
    \caption{Obesity}
    \label{fig:true_obesity}
\end{subfigure}

\begin{subfigure}{0.3\textwidth}
    \includegraphics[width=\linewidth]{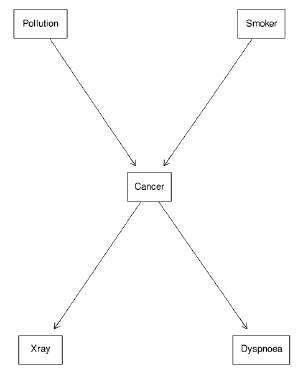}
    \caption{Cancer}
    \label{fig:true_cancer}
\end{subfigure}
\hfill
\begin{subfigure}{0.5\textwidth}
    \includegraphics[width=\linewidth]{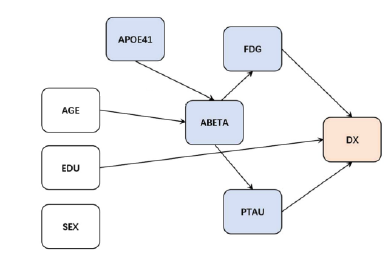}
    \caption{Alzheimer’s Disease Neuroimaging Initiative (ADNI)}
    \label{fig:true_adni}
\end{subfigure}

\begin{subfigure}{0.8\textwidth}
    \includegraphics[width=\linewidth]{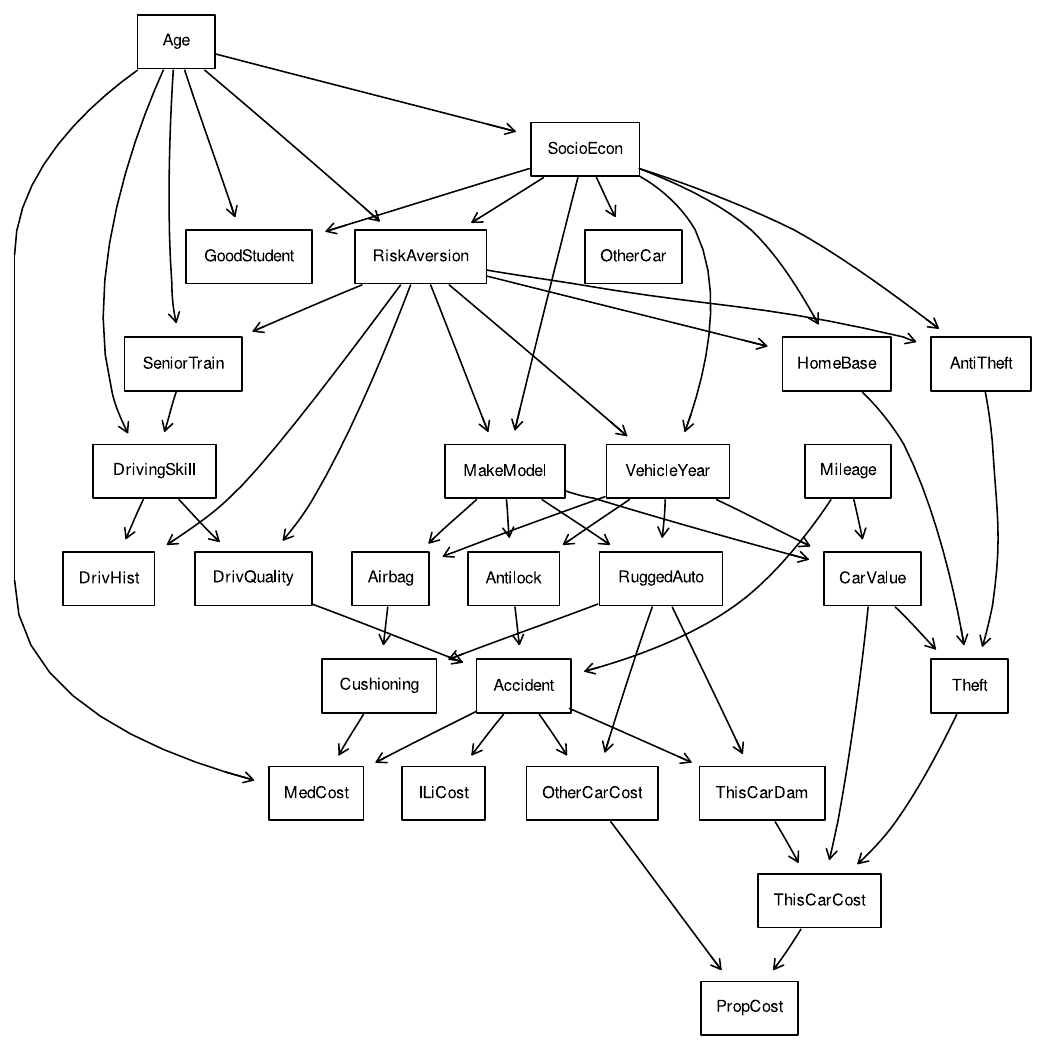}
    \caption{Insurance}
    \label{fig:true_insurance}
\end{subfigure}

\caption{The ground-truth causal graphs from original sources  \cite{hernan2004structural, long2022can, shen2020challenges, 10.5555/1941985, binder1997adaptive}.}
\label{fig:ground_truth_causal_graphs}
\end{figure*}

The ground-truth causal graphs of initial variables for evaluating the causal discovery component can be found in \Figref{fig:ground_truth_causal_graphs}. Table~\ref{tab:statistics_initial_graphs} demonstrates the statistics of initial ground-truth causal graphs with initial variables.

\begin{table}[t]
\centering
% \resizebox{\linewidth}{!}{%
\begin{tabular}{l|cc}
\hline
Causal Graph        & Node & Edge \\ \hline
Cancer              & 5    & 4    \\
Respiratory Disease & 4    & 5    \\
Diabetes            & 4    & 5    \\
Obesity             & 4    & 5    \\
ADNI                & 8    & 7    \\
Insurance           & 27   & 52   \\ \hline
\end{tabular}%
% }
\caption{Statistics of ground-truth causal graph for initial variables.}
\label{tab:statistics_initial_graphs}
\end{table}

\subsection{Evaluation of Causal Discovery Component}
\label{apx:eval_causal_discovery_component}

The detailed evaluation results of the causal discovery component are presented in Table~\ref{tab:result_causal_discovery_cancer},~\ref{tab:result_causal_discovery_respiratory_disease}, ~\ref{tab:result_causal_discovery_diabetes}, ~\ref{tab:result_causal_discovery_obesity}, ~\ref{tab:result_causal_discovery_adni}, and ~\ref{tab:result_causal_discovery_insurance}.

\begin{table*}[t]
\centering
% \resizebox{\textwidth}{!}{%
\begin{tabular}{lccccc}
\hline
\multicolumn{6}{c}{Cancer (5 nodes, 4 edges)}                                                                       \\
Method                           & Precision & Recall & \textbf{F1↑}  & \# of predicted edges & \textbf{NHD Ratio↓} \\ \hline
Pairwise-LLM                     & 0.75      & 0.75   & 0.75          & 4                     & 0.25                \\
BFS-LLM                          & 0.6       & 0.75   & 0.67          & 5                     & 0.33                \\
COAT                             & 0.13      & 0.25   & 0.17          & 8                     & 0.83                \\
\ourmethod - GES                 & 0.25      & 0.5    & 0.33          & 8                     & 0.67                \\
\ourmethod - NOTEARS             & 1.0       & 0.25   & 0.4           & 1                     & 0.6                 \\
\ourmethod  - PC                 & 0.14      & 0.25   & 0.18          & 7                     & 0.82                \\
\ourmethod - VCR                 & 1.0       & 0.75   & \textbf{0.86}          & 3                     & \textbf{0.14}                \\
\ourmethod (Llama) - NOTEARS+VCR & 0.375     & 0.75   & 0.5           & 8                     & 0.5                 \\
\ourmethod  - NOTEARS+VCR        & 1.0       & 0.75   & \textbf{0.86} & 3                     & \textbf{0.14}       \\ \hline
\end{tabular}%
% }
\caption{Evaluation results of causal discovery on cancer graph. VCR refers to verified causal relations that are extracted from and validated by relevant academic documents. "Llama" refers to the use of the Llama-3.1-8b-instruct model as a substitute for GPT-4o in our method.}
\label{tab:result_causal_discovery_cancer}
% \vspace{-10pt}
\end{table*}
\begin{table*}[t]
\centering
% \resizebox{\textwidth}{!}{%
\begin{tabular}{lccccc}
\hline
\multicolumn{6}{c}{Respiratory Disease (4 nodes, 5 edges)}                                                     \\
Method                      & Precision & Recall & \textbf{F1↑}  & \# of predicted edges & \textbf{NHD Ratio↓} \\ \hline
Pairwise-LLM                & 1.0       & 0.6    & 0.75          & 3                     & 0.25                \\
BFS-LLM                     & 0.67      & 0.4    & 0.5           & 3                     & 0.5                 \\
COAT                        & 1.0       & 0.8    & 0.89          & 4                     & 0.11                \\
\ourmethod - GES            & 1.0       & 0.8    & 0.89          & 4                     & 0.11                \\
\ourmethod - NOTEARS        & 1.0       & 0.2    & 0.33          & 1                     & 0.67                \\
\ourmethod - PC             & 0.83      & 1.0    & 0.91          & 6                     & 0.09                \\
\ourmethod - VCR            & 1.0       & 0.8    & 0.89          & 4                     & 0.11                \\
\ourmethod (Llama) - PC+VCR & 1.0       & 0.8    & 0.89          & 4                     & 0.11                \\
\ourmethod - PC+VCR         & 0.83      & 1.0    & \textbf{0.91} & 6                     & \textbf{0.09}       \\ \hline
\end{tabular}%
% }
\caption{Evaluation results of causal discovery on respiratory disease graph. }
\label{tab:result_causal_discovery_respiratory_disease}
% \vspace{}
\end{table*}

\begin{table*}[t]
\centering
% \resizebox{\textwidth}{!}{%
\begin{tabular}{lccccc}
\hline
\multicolumn{6}{c}{Diabetes (4 nodes, 5 edges)}                                                                 \\
Method                       & Precision & Recall & \textbf{F1↑}  & \# of predicted edges & \textbf{NHD Ratio↓} \\ \hline
Pairwise-LLM                 & 0.67      & 0.4    & 0.5           & 3                     & 0.5                 \\
BFS-LLM                      & 0.67      & 0.4    & 0.5           & 3                     & 0.5                 \\
COAT                         & 0.25      & 0.2    & 0.22          & 4                     & 0.78                \\
\ourmethod - GES             & 0.5       & 0.6    & 0.55          & 6                     & 0.45                \\
\ourmethod - NOTEARS         & 0         & 0      & 0             & 0                     & 1.0                 \\
\ourmethod - PC              & 0.25      & 0.2    & 0.22          & 4                     & 0.78                \\
\ourmethod - VCR             & 1.0       & 0.2    & 0.33          & 1                     & 0.67                \\
\ourmethod (Llama) - GES+VCR & 0.67      & 0.4    & 0.5           & 3                     & 0.5                 \\
\ourmethod - GES+VCR         & 1.0       & 0.6    & \textbf{0.75} & 3                     & \textbf{0.25}       \\ \hline
\end{tabular}%
% }
\caption{Evaluation results of causal discovery on diabetes graph.}
\label{tab:result_causal_discovery_diabetes}
\end{table*}

\begin{table*}[t]
\centering
% \resizebox{\textwidth}{!}{%
\begin{tabular}{lccccc}
\hline
\multicolumn{6}{c}{Obesity (4 nodes, 5 edges)}                                                                \\
                            & Precision & Recall & \textbf{F1↑} & \# of predicted edges & \textbf{NHD Ratio↓} \\ \hline
Pairwise-LLM                & 0.83      & 1.0    & 0.91         & 6                     & 0.09                \\
BFS-LLM                     & 0.6       & 0.6    & 0.6          & 5                     & 0.4                 \\
COAT                        & 0.25      & 0.2    & 0.22         & 4                     & 0.78                \\
\ourmethod -GES             & 0.25      & 0.2    & 0.22         & 4                     & 0.78                \\
\ourmethod - NOTEARS        & 0         & 0      & 0            & 2                     & 1.0                 \\
\ourmethod - PC             & 0.25      & 0.2    & 0.22         & 4                     & 0.78                \\
\ourmethod - VCR            & 1.0       & 1.0    & \textbf{1.0} & 5                     & \textbf{0}          \\
\ourmethod (Llama) - PC+VCR & 0.83      & 1.0    & 0.91         & 6                     & 0.09                \\
\ourmethod - PC+VCR         & 1.0       & 1.0    & \textbf{1.0} & 5                     & \textbf{0}          \\ \hline
\end{tabular}%
% }
\caption{Evaluation results of causal discovery on obesity graph.}
\label{tab:result_causal_discovery_obesity}
% \vspace{-10pt}

\end{table*}

\begin{table*}[t]
\centering
% \resizebox{\textwidth}{!}{%
\begin{tabular}{lccccc}
\hline
\multicolumn{6}{c}{ADNI (8 nodes, 7 edges)}                                                                        \\
Method                           & Precision & Recall & \textbf{F1↑} & \# of predicted edges & \textbf{NHD Ratio↓} \\ \hline
Pairwise-LLM                     & 0.5       & 0.14   & 0.22         & 2                     & 0.78                \\
BFS-LLM                          & 0.33      & 0.14   & 0.2          & 3                     & 0.8                 \\
COAT                             & 0.11      & 0.14   & 0.13         & 9                     & 0.87                \\
\ourmethod - GES                 & 0.08      & 0.14   & 0.11         & 12                    & 0.89                \\
\ourmethod - NOTEARS             & 0.33      & 0.14   & 0.2          & 3                     & 0.8                 \\
\ourmethod - PC                  & 0.11      & 0.14   & 0.13         & 9                     & 0.87                \\
\ourmethod - VCR                 & 0.4       & 0.29   & 0.33         & 5                     & 0.67                \\
\ourmethod (Llama) - NOTEARS+VCR & 0.08      & 0.14   & 0.11         & 12                    & 0.89                \\
\ourmethod - NOTEARS+VCR         & 0.38      & 0.43   & \textbf{0.4} & 8                     & \textbf{0.6}        \\ \hline
\end{tabular}%
% }
\caption{Evaluation results of causal discovery on ADNI graph.}
\label{tab:result_causal_discovery_adni}
\vspace{-10pt}
\end{table*}

\begin{table*}[t]
\centering
% \resizebox{\textwidth}{!}{%
\begin{tabular}{lccccc}
\hline
\multicolumn{6}{c}{Insurance (27 nodes, 52 edges)}                                                              \\
Method                       & Precision & Recall & \textbf{F1↑}  & \# of predicted edges & \textbf{NHD Ratio↓} \\ \hline
Pairwise-LLM                 & 0.37      & 0.34   & 0.35          & 58                    & 0.68                \\
BFS-LLM                      & 0.33      & 0.24   & 0.28          & 41                    & 0.76                \\
COAT                         & 0.38      & 0.37   & 0.37          & 53                    & 0.65                \\
\ourmethod - GES             & 0.41      & 0.47   & 0.44          & 37                    & 0.58                \\
\ourmethod - NOTEARS         & 0.28      & 0.40   & 0.32          & 33                    & 0.71                \\
\ourmethod - PC              & 0.31      & 0.40   & 0.35          & 31                    & 0.69                \\
\ourmethod - VCR             & 0.4       & 0.29   & 0.33          & 35                    & 0.67                \\
\ourmethod (Llama) - GES+VCR & 0.43      & 0.49   & 0.46          & 43                    & 0.55                \\
\ourmethod - GES+VCR         & 0.58      & 0.53   & \textbf{0.55} & 47                    & \textbf{0.45}       \\ \hline
\end{tabular}%
% }
\caption{Evaluation results of causal discovery on Insurance graph.}
\label{tab:result_causal_discovery_insurance}
\end{table*}

\end{document}